\documentclass{article}

\PassOptionsToPackage{numbers, compress}{natbib} % 숫자 레퍼런스 사용을 위함

\usepackage[preprint]{neurips_2025}

\usepackage[utf8]{inputenc} % allow utf-8 input
\usepackage[T1]{fontenc}    % use 8-bit T1 fonts
\usepackage{hyperref}       % hyperlinks
\usepackage{url}            % simple URL typesetting
\usepackage{booktabs}       % professional-quality tables
\usepackage{amsfonts}       % blackboard math symbols
\usepackage{nicefrac}       % compact symbols for 1/2, etc.
\usepackage{microtype}      % microtypography
\usepackage{xcolor}         % colors

\usepackage{mhchem}
\usepackage{graphicx}
\usepackage{animate}

\usepackage{cuted}
\usepackage{tabularx}
\usepackage{capt-of}

\usepackage{multirow}
\usepackage{multicol}
\usepackage{titletoc}

\usepackage{wrapfig} % 이미지나 표 주위에 텍스트가 감싸도록
\usepackage{xcolor}   % 색상 패키지 로드, 글씨 색깔 바꾸기용
\usepackage{cleveref}  % cref 사용 위함
\usepackage{caption}  % wrapfigure시 텍스트가 그림을 침범하지 못하게 하기 위해 captionsetup을 사용할 때 필요

\usepackage{amsmath, amssymb}
\usepackage{algorithm}
\usepackage[noend]{algpseudocode}
\usepackage[most]{tcolorbox}
\tcbuselibrary{listings, breakable}

\definecolor{darkgreen}{rgb}{0.0, 0.5, 0.0}
\definecolor{darkyellow}{rgb}{0.9, 0.9, 0.0}
\newcommand{\modelname}{TV-LiVE}
\title{\modelname: Training-Free, Text-Guided Video Editing \\
via Layer Informed Vitality Exploitation}
\author{
  Min-Jung Kim\textsuperscript{1},
  Dongjin Kim\textsuperscript{1},
  Seokju Yun\textsuperscript{2},
  Jaegul Choo\textsuperscript{1} \\
  \textsuperscript{1}KAIST AI \quad
  \textsuperscript{2}University of Seoul \\
  \texttt{\{emjay73, dj\_kim, jchoo\}@kaist.ac.kr, wsz871@uos.ac.kr}
}

\begin{document}

\maketitle

\begin{center}
% \begin{teaserfigure}
    \newcommand{\numColumns}{4}
    \newcommand{\columnSpacing}{0.25em}

    \begin{tabular}{
        @{}
        p{\dimexpr(\textwidth-\columnSpacing*(\numColumns-1))/\numColumns} @{\hspace{\columnSpacing}}
        p{\dimexpr(\textwidth-\columnSpacing*(\numColumns-1))/\numColumns} 
        @{\hspace{\columnSpacing}}
        p{\dimexpr(\textwidth-\columnSpacing*(\numColumns-1))/\numColumns} 
        @{\hspace{\columnSpacing}}
        p{\dimexpr(\textwidth-\columnSpacing*(\numColumns-1))/\numColumns} 
        @{}
    }
        \animategraphics[loop, width=\linewidth]{8}{videos/small_teaser1_gardening_gloves/}{0000}{48} &
        \animategraphics[loop, width=\linewidth]{8}{videos/small_teaser2_turtle_leaf/}{0000}{48} &
        \animategraphics[loop, width=\linewidth]{8}{videos/small_teaser3_hummingbird/}{0000}{48} &
        \animategraphics[loop, width=\linewidth]{8}{videos/small_teaser4_wolf_howling/}{0000}{48}
    \end{tabular}
    \begin{tabularx}{\textwidth}{XXXX}
        \centering {\scriptsize\selectfont "+ Gardening Gloves"} & 
        \centering {\scriptsize\selectfont "+ Leaf"} & 
        \centering {\scriptsize\selectfont "+ Extending its beak \\into the flower"} & 
        \centering {\scriptsize\selectfont "+ Howling with its head \\tilted upward"}
    \end{tabularx}

    \captionof{figure}{Video editing results conducted using our \modelname. The left two columns demonstrate video editing with object addition, and the right two columns show video editing with non-rigid deformation. Best viewed with Acrobat Reader. Click each image to play the video clip.}
    \label{fig:main}
\end{center}

\begin{abstract}
  % Two line spaces precede the abstract. 
  % The abstract must be limited to one paragraph.
  Video editing has garnered increasing attention alongside the rapid progress of diffusion-based video generation models. 
  As part of these advancements, there is a growing demand for more accessible and controllable forms of video editing, such as prompt-based editing. 
  Previous studies have primarily focused on tasks such as style transfer, background replacement, object substitution, and attribute modification, while maintaining the content structure of the source video. However, more complex tasks, including the addition of novel objects and non-rigid transformations, remain relatively unexplored.
  In this paper, we present \modelname{}, a \textit{T}raining-free and text-guided \textit{V}ideo editing framework via \textit{L}ayer-\textit{i}nformed \textit{V}itality \textit{E}xploitation. We empirically identify vital layers within the video generation model that significantly influence the quality of generated outputs. Notably, these layers are closely associated with Rotary Position Embeddings (RoPE).
  Based on this observation, our method enables both object addition and non-rigid video editing by selectively injecting key and value features from the source model into the corresponding layers of the target model guided by the layer vitality. For object addition, we further identify prominent layers to extract the mask regions corresponding to the newly added target prompt. We found that the extracted masks from the prominent layers faithfully indicate the region to be edited.
  Experimental results demonstrate that \modelname{} outperforms existing approaches for both object addition and non-rigid video editing.
  %
  % Code will be released upon acceptance.
  Project Page: \url{https://emjay73.github.io/TV_LiVE/}
  
\end{abstract}

\section{Introduction}
% 최근 비디오 생성 모델의 발전이 빠르게 이루어짐( 구조적으로도 unet 에서 dit로의 변화가 이루어짐) 
% 비디오 편집 모델 연구의 필요성도 함께 발전 
% 비디오 편집이 가능해지면 산업계에서 좋은 점(아주 간략)
% 하지만 최신 비디오 편집기술들은 물체를 추가하거나 객체를 non-rigid하게 변형하는 것이 불가능.
% 우리 연구에서는 이 두가지를 가능하게 하는 training free 비디오 생성 연구를 수행함
% contribution
% - unet 기반 연구와 한계 소개?
% - dit 기반 연구와 한계 소개?

Recent progress in video generation has been driven by the evolution of model architectures, shifting from convolutional U-Net-based designs~\cite{blattmann2023align, wang2023modelscope,guo2023animatediff} to more expressive transformer-based structures such as the Denoising Diffusion Transformer (DiT)~\cite{hong2022cogvideo, gupta2024walt, yang2024cogvideox, kong2024hunyuanvideo}. These advancements have significantly improved the quality and coherence of generated videos, opening the door to a wide range of downstream applications~\cite{wen2024panacea, xing2024survey}. Among them, video editing has emerged as a critical yet less explored challenge.

In practical settings, editable video generation offers strong advantages: enabling content creation~\cite{wu2023tune, ceylan2023pix2video}, reducing production costs~\cite{singer2022make, khachatryan2023text2video, qi2023fatezero}, and supporting interactive media workflows~\cite{gao2024assisteditor}. However, despite recent efforts, state-of-the-art video editing models~\cite{shi2024bivdiff, kara2024rave, li2024vidtome} remain limited in scope. Most are constrained to simple appearance changes or rigid geometric modifications. They are unable to handle more complex editing tasks such as introducing new objects into a scene or applying non-rigid semantic transformations (e.g., changing human pose or expression), both of which are crucial for real-world applicability.

In this work, we present a training-free, text-guided video editing framework built upon DiT-based video generation models. Our approach allows both object insertion and non-rigid semantic edits without the need for model fine-tuning or additional training, making it highly efficient and flexible. To better understand the internal behavior of DiT models in the context of video editing, we conduct a layer-wise analysis using two metrics, vitality and prominence, which help identify layer-specific roles in video generation control. These insights can inform future research on video-conditioned generation and model interpretability.

Empirically, our method achieves superior performance on standard video editing benchmarks, surpassing existing approaches in both qualitative fidelity and quantitative accuracy. Our results suggest that even without training, DiT-based video generators can be effectively repurposed for complex editing tasks when guided by appropriate structural understanding and prompt-level control.

To summarize, our contributions are as follows: (1) We propose a training-free, text-guided video editing method for DiT-based video generation models, (2) We analyze the internal layer properties of DiT-based models using two metrics: vitality and prominence. This analysis provides insights into the design of video editing applications and other related tasks, and (3) Our method outperforms recent video editing approaches both qualitatively and quantitatively.

% \setlength\itemindent{0em} % 1em: 글꼴 크기 단위
% \begin{enumerate}
%     \item We propose a training-free, text-guided video editing method for DiT-based video generation models.
%     \item We analyze the internal layer properties of DiT-based models using two metrics: vitality and prominence. This analysis provides insights into the design of video editing applications and other related tasks.
%     \item Our method outperforms recent video editing approaches both qualitatively and quantitatively.
% \end{enumerate}

\section{Related Work}
\subsection{T2V Video Generation Model}
% unet 기반 video model (animate diff, align your latent..) 을 거쳐 dit모델(cogvideo, hunyuan)로 넘어가는 추세이다.
Early text-to-video generation models build upon U-Net architectures. Blattmann \textit{et al.} \cite{blattmann2023align} train image models first, then introduce temporal layers during fine-tuning to achieve high-resolution results. ModelScopeT2V~\cite{wang2023modelscope} and AnimateDiff~\cite{guo2023animatediff} follow similar approaches, but these U-Net based models exhibit limitations with complex motion and longer durations.
Recent approaches shift toward Diffusion Transformer (DiT) architectures for improved long-range modeling and scalability ~\cite{hong2022cogvideo, yang2024cogvideox, gupta2024walt, kong2024hunyuanvideo}. CogVideo~\cite{hong2022cogvideo} employs a 9B-parameter transformer, while W.A.L.T.~\cite{gupta2024walt} demonstrates the superiority of transformers over U-Nets in video diffusion models. CogVideoX~\cite{yang2024cogvideox} generates 10-second videos (16fps, 768×1360), and HunyuanVideo~\cite{kong2024hunyuanvideo} utilizes a 13B-parameter DiT backbone to enhance text-video alignment. Transformer-based models capture temporal dependencies more effectively, substantially improving the quality and realism of generated videos.

\subsection{Diffusion Based Image Editing}
Diffusion models revolutionize image editing approaches. SDEdit~\cite{meng2021sdedit} demonstrates that controlled noise addition and denoising preserves structure while enabling edits. DiffEdit~\cite{couairon2022diffedit} introduces automatic masking, while Prompt-to-Prompt~\cite{hertz2022prompt} manipulates cross-attention layers for targeted modifications. Inversion methods like Null-text inversion~\cite{mokady2023null} and instruction-based approaches like InstructPix2Pix~\cite{brooks2023instructpix2pix} improve source fidelity. Recent advances include FreeFlux~\cite{wei2025freeflux}, which categorizes editing tasks with tailored key-value injection strategies, and Stable Flow~\cite{avrahami2024stable}, which identifies vital layers for controlled edits in flow-based models. While these methods achieve impressive results for single images, extending such techniques to the temporal domain introduces unique challenges that require specialized approaches to maintain consistency across video frames.

\subsection{Diffusion Based Video Editing}
Tune-A-Video~\cite{wu2023tune} fine-tunes on single videos to preserve motion, while training-free approaches include FateZero~\cite{qi2023fatezero}, which leverages attention maps to encode structural information, and Pix2Video~\cite{ceylan2023pix2video}, which propagates edits through modified self-attention layers. CogVideoX \cite{yang2024cogvideox} introduces CogInv (DDIM inversion-based) and CogV2V (SDEdit-based) components for text-guided editing with temporal consistency. Advanced techniques include Video-P2P's \cite{liu2024video} cross-frame attention mechanisms, Dreamix's \cite{molad2023dreamix} motion-aware conditioning, TokenFlow's \cite{geyer2023tokenflow} feature correspondence propagation, and various approaches like BIVDiff \cite{shi2024bivdiff}, RAVE \cite{kara2024rave}, and VidToMe \cite{li2024vidtome} that enhance editing capabilities through novel frameworks.

While existing methods excel at style transfer, background replacement, and attribute modification, they struggle with text-based object addition and non-rigid transformations. In this paper, we propose training-free, text-guided video editing via layer informed vitality exploitation specifically targets these limitations by leveraging dynamic motion information at multiple feature layers of the diffusion model. This approach enables both rigid and non-rigid editing while preserving static background content and focusing modifications on intended moving subjects.

\section{Preliminary}
\noindent \textbf{CogVideoX}
CogVideoX~\cite{yang2024cogvideox} is a widely adopted DiT-based text-to-video generation model.
% First it downsamples video into latent space. After 8x downsampling compression in spatial dimensions and 4x downsampling compression in temporal dimensions, and additional 2x patchfication in spatial dimensions, the latent has the sequential length that is $1\over{64}$ compared to its original size.
% For instance, when the width of video is 720, height is 480 and temporal length is 49, video in latent space would have a sequential length of 45x30x13.
The architecture consists solely of self-attention blocks, without any cross-attention layers. The text prompt embeddings (length 226) and visual latent embeddings are concatenated along the sequence dimension and jointly fed into the self-attention mechanism as follows:
\begin{equation}
    Attn = softmax([Q_{txt}, RoPE(Q_{vis})][K_{txt}, RoPE(K_{vis})]^T/\sqrt{d})[V_{txt}, V_{vis}],
\end{equation}
where $[\cdot, \cdot]$ denotes concatenation and $\text{RoPE}(\cdot)$ denotes the application of Rotary Positional Embedding, and where $Q_{\text{txt}}, K_{\text{txt}}, V_{\text{txt}}$ represent the query, key, and value obtained from the text prompt embeddings, while $Q_{\text{vis}}, K_{\text{vis}}, V_{\text{vis}}$ denote those derived from the visual latent representations of the video.
In the following section, we explore how to inject the key and value tensors and into which layers of the target model to inject them.
% In the following section, the key and value representations of the visual components will be replaced with externally injected representations at specific layers. 
Unless otherwise stated, any reference to "key" and "value" pertains solely to the \textit{visual} components.
% 
% \subsection{RoPE embedding}
% % In this paper, we introduce training-free text-guided video editing method using DiT based text-to-video diffusion model.

\section{Method}
We consider two video editing objectives guided by a text prompt.  
The first is \textit{object addition}, where new objects are introduced into the video while preserving the original, unedited regions.  
The second is \textit{non-rigid editing}, where the motion or pose of existing elements is modified, but the overall content and appearance of the source video are retained.
Formally, given a source prompt $p_s$ describing the input video and a target prompt $p_t$ describing the intended edit, the goal is to generate an edited video $\hat{\mathbf{V}}$ that reflects the changes in $p_t$ while maintaining consistency with the original video, either by preserving background regions (object addition) or maintaining overall appearance and identity (non-rigid editing).

To achieve the aforementioned goals in a training-free setting, we analyze each layer of the model to identify which ones are most relevant to the editing task.
In \Cref{sec:vitality}, we quantify the relative importance of each layer through a vitality analysis, where we either bypass the layer entirely or drop the Rotary Positional Embedding (RoPE) from its key projections.
Based on this analysis, we perform object addition using the layers identified as vital in \Cref{sec:obj_add}, and non-rigid editing using the non-vital layers in \Cref{sec:non_rigid}.

\begin{figure}[t]
    \centering    
    \includegraphics[width=\linewidth]{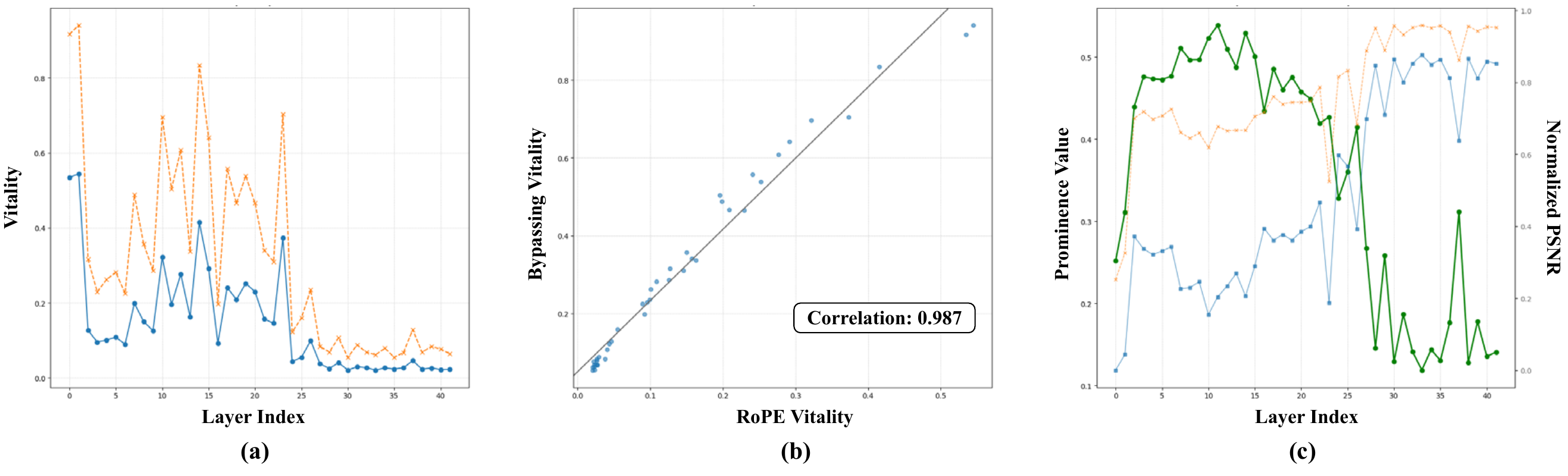}
    \caption{\textbf{Layer Vitality and Layer Prominence Analysis}. 
    \textbf{(a)} \textcolor{orange}{The orange graph} shows Bypassing Vitality, measured by bypassing each layer, while \textcolor{blue}{the blue graph} shows RoPE Vitality, measured by removing RoPE from the key of each attention layer. 
    \textbf{(b)} The Pearson correlation coefficient between Bypassing Vitality and RoPE Vitality is 0.987, indicating a strong correlation between the two metrics. 
    \textbf{(c)} \textcolor{darkgreen}{The green graph} represents the prominence value of individual layers, computed according to~\Cref{eq:prominence}. \textcolor{orange}{The orange} and \textcolor{blue}{blue graphs} depict the normalized PSNR between the original and RoPE-dropped outputs, evaluated separately over the background and foreground regions. 
}
    \label{fig:layer_analysis}
\end{figure}

\subsection{Layer Vitality Analysis}
\label{sec:vitality}
% layer 특성을 아는 것의 중요성
Understanding the functional role of each layer is crucial in video generation models, particularly when deciding which layers to train or modify for downstream tasks. Recent methods have effectively leveraged well-established properties of the U-Net architecture to align model behavior with specific goals. For example, P+~\cite{voynov2023PP} shows that the coarse layers in U-Net primarily govern structural layout, while fine layers are responsible for refining appearance.
Building on this understanding, several video generation methods based on image diffusion models~\cite{kim2024tcan, xu2024magicanimate, nguyen2025swifttry, zheng2025soyo} incorporate the appearance of source media by injecting or concatenating key/value features, which are typically applied at the fine-resolution decoder layers.
% 
% 어떻게 layer의 중요도를 판단할 것인가
%% 두 연구(stable flow, freeflux)의 영향을 받아 두가지 방법을 시도 
However, layer-wise analysis remains relatively underexplored in DiT-based video generation models. Motivated by recent approaches~\cite{avrahami2024stable, wei2025freeflux, }, we evaluate the \textit{layer vitality} of DiT-based video generators using two complementary techniques: (1) bypassing each layer individually, and (2) removing Rotary Positional Embedding (RoPE) from the key projection of each layer.

% 
%% 1. bypassing
% 
We first use ChatGPT~\cite{openai2023chatgpt} to generate a set $S$ of $N_p$ diverse text prompts 
and produce corresponding videos via CogVideoX~\cite{yang2024cogvideox}.  
% Using each prompt in $S$, we generate corresponding videos via a CogVideoX~\cite{yang2024cogvideox}.
For each video, we create a series of probing videos by bypassing each of the $42$ layers individually while keeping the text prompt and random seed fixed. 
Specifically, we bypass a layer by 
% feeding the output of the previous layer directly into the next, effectively skipping the computation of that layer.
skipping its computation and forwarding the previous layer’s output.
After generating all probing videos, 
% we compute the perceptual similarity between each probing video and its corresponding original video using DINOv2~\cite{oquab2023dinov2}.  
% We then formally define the vitality of a layer $l$ as follows:
we measure perceptual similarity to the original using DINOv2~\cite{oquab2023dinov2} and define the vitality of layer $l$ as:

\begin{equation}
    % vitality^{layer}(l) = 1-{1\over{N_p}}\sum_{s\in S} {sim(Dinov2(V_o), Dinov2(V^{layer}_l))},
    \text{vitality}^{\text{layer}}(l) = 1 - \frac{1}{N_p} \sum_{s \in S} \text{sim}(\text{DINOv2}(V_o), \text{DINOv2}(V^{\text{layer}}_l)),
\end{equation}
where $V_o$ is the original (unmodified) video, $V^{\text{layer}}_l$ is the probing video obtained by bypassing layer $l$, and $sim(\cdot, \cdot)$ denotes the perceptual cosine similarity between DINOv2 features.

%% 2. RoPE dropping 
In a similar manner, we compute an alternative vitality score by removing the RoPE from the attention key of each layer.  
Let $V^{\text{RoPE}}_l$ denote the probing video generated by dropping RoPE from layer $l$.  
The RoPE-based vitality is defined as:
\begin{equation}
    \text{vitality}^{\text{RoPE}}(l) = 1 - \frac{1}{N_p} \sum_{s \in S} \text{sim}(\text{DINOv2}(V_o), \text{DINOv2}(V^{\text{RoPE}}_l)),
\end{equation}
where $V^{\text{RoPE}}_l$ is specific to layer $l$ and captures the impact of RoPE removal on the output.

%%% RoPE의 중요성을 언급했던 논문들 ref (freeflux포함)??
%% 둘이 매우 높은 유사도 보임. RoPE의 위치 의존도가 높은 layer들이 vital layer로 간주됨.
As shown in~\Cref{fig:layer_analysis}(a) and~\Cref{fig:layer_analysis}(b), we observe a strong correlation between layer vitality scores obtained via the layer bypass method and those obtained through RoPE removal. This correlation suggests that the removal of RoPE from certain layers can induce effects on video generation comparable to those caused by bypassing the entire layer.
Motivated by this observation, we hypothesize that \textit{layers exhibiting sensitivity to positional encoding (i.e., position-dependent layers) are more likely to be vital for model performance, whereas position-independent layers tend to be non-vital.}
Throughout this paper, we define vital layers as those exhibiting high vitality scores when RoPE is removed, indicating a strong dependency on positional information.

% 비디오 editing연구에 어떻게 활용할 것인가
%% stableflow처럼 vital layer에 key/value injection 수행한 결과 소스 비디오를 따라갈 뿐 object addition과 non-rigid editing을 수행하는 것은 불가능
The image editing method StableFlow~\cite{avrahami2024stable} demonstrates that both object addition and non-rigid transformations can be achieved by injecting keys and values into the vital layers identified via layer bypassing.
However, in our experiments, directly applying this approach fails to yield the intended effects: the generated video remains identical to the source regardless of the target prompt, indicating a lack of controllability.
To address this limitation, we adopt an alternative strategy.

% Similar to FreeFlux~\cite{wei2025freeflux}, 
For the object addition task, we inject the keys and values of the vital layers from the source video into the corresponding layers of the target video. This encourages the model to follow the spatial structure of the source video in a position-wise manner. 
To accommodate newly introduced objects, we adopt a masked key-value injection strategy, similar to the method in
~\cite{cao2023masactrl, hertz2022prompt}, 
where keys and values are not injected within the masked region. This prevents the masked area from inheriting position-specific information from the source, thereby enabling the model to reflect the target prompt in that region.
For non-rigid deformation, we instead inject keys and values into non-vital layers, i.e., layers that exhibit lower sensitivity to positional encoding, so as to transfer the appearance information from the source video without enforcing strict spatial alignment.
%% 위치 의존도가 높은 vital layer는 object addition에, 위치 의존도가 낮은 non-vital layer는 non-rigid editing에 적합할 것 이라는 가정을 가지고 접근.
\subsection{Video Editing for Object Addition}
\label{sec:obj_add}
\begin{figure}[t]
    \centering      
    % \includegraphics[width=0.9\linewidth]{figures/main_method.pdf}
    % \caption{Overview of our object addition framework. Up to the denoising timestep $T_i$, standard key/value injection is applied to vital layers. From $T_i$ to $T_e$, masked key/value injection is performed. The mask is obtained by averaging and then binarizing the attention maps of the prominent layer over the last three timesteps during the standard injection phase.}
    % \label{fig:main_method}
    \includegraphics[width=0.9\linewidth]{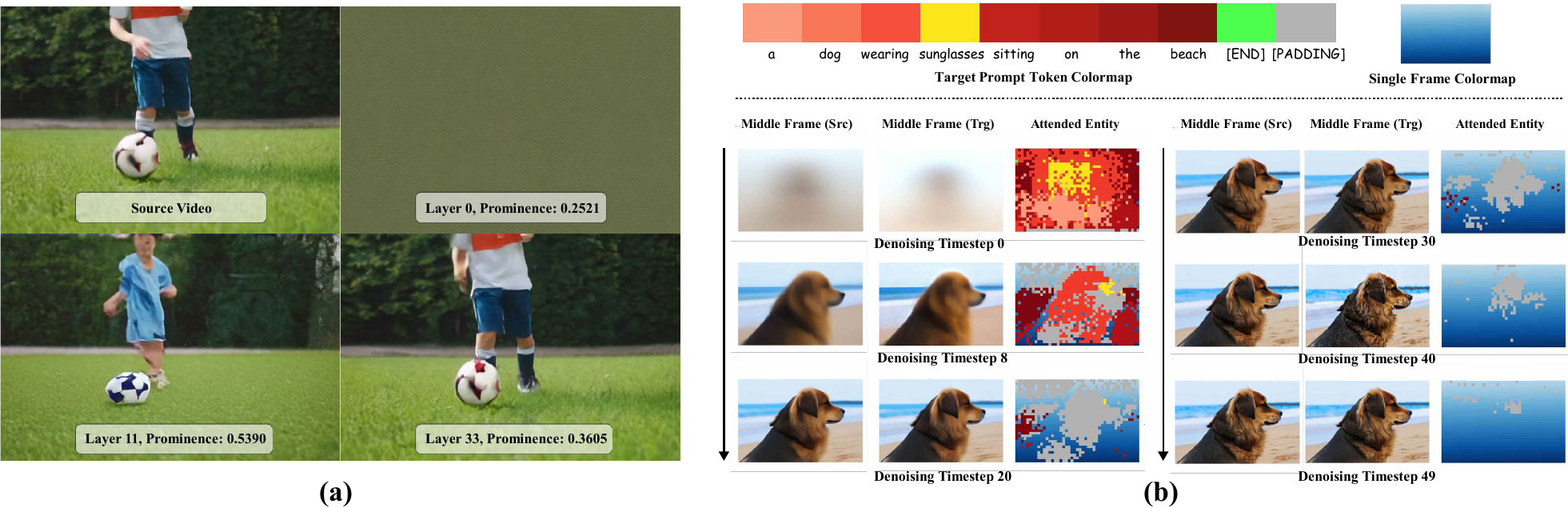}
    % \caption{\textbf{Layer Prominence Visualization}. (a) The top-left image is from the source video, and the others show results when RoPE is ablated at each layer. Layer 11, which mainly affects the foreground, has the highest prominence score. (b) Visualization of where the target video attend the most at each denoising timestep, when the key and value of vital layers are injected without mask. Red represents target prompt, yellow represents delta token, and blue represents source video. Around timestep 8, we observed that the target video has finished structuring the video contents and start to attend to delta token (\textcolor{darkyellow}{Yellow}) at proper location. Only middle frame is visualized for the simplicity.}  % 이미지 캡션
    \caption{\textbf{Layer Prominence Visualization}. (a) The top-left image is from the source video, and the others show results when RoPE is ablated at each layer. Layer 11, which mainly affects the foreground, has the highest prominence score. (b) Visualization of where the target video attends the most at each denoising timestep. Key and value features are injected without a mask. \textcolor{red}{Red}: target prompt, \textcolor{yellow}{Yellow}: delta token, \textcolor{blue}{Blue}: source video. Around timestep 8, the target video starts attending to the delta token at the correct location. Only the middle frame is shown for clarity.}
    \label{fig:prominent_layer}
\end{figure}
% Directly injecting keys and values into the vital layers causes the target video to reconstruct the source video almost identically, leaving little room for modification. 
To select the most suitable layer for mask extraction in the masked key-value injection framework, we first analyze which layers are most influential in shaping the foreground content. 
% This is done by identifying foreground regions in each frame, guided by the subject term in the prompt. 
% This process utilizes Grounded Sam2~\cite{ren2024grounded}, which integrates Grounding DINO~\cite{ren2024grounded} with SAM 2~\cite{ravi2024sam2segmentimages}.
% We then compute the PSNR between the sampled video and a video generated by dropping RoPE at a specific layer, averaging the values over both spatial and temporal dimensions.
Our goal is to identify layers that have minimal influence on the background while significantly impacting the foreground region.

\noindent \textbf{Layer Prominence Analysis}
\begin{figure}[t!]
    \centering
    \includegraphics[width=0.8\linewidth]{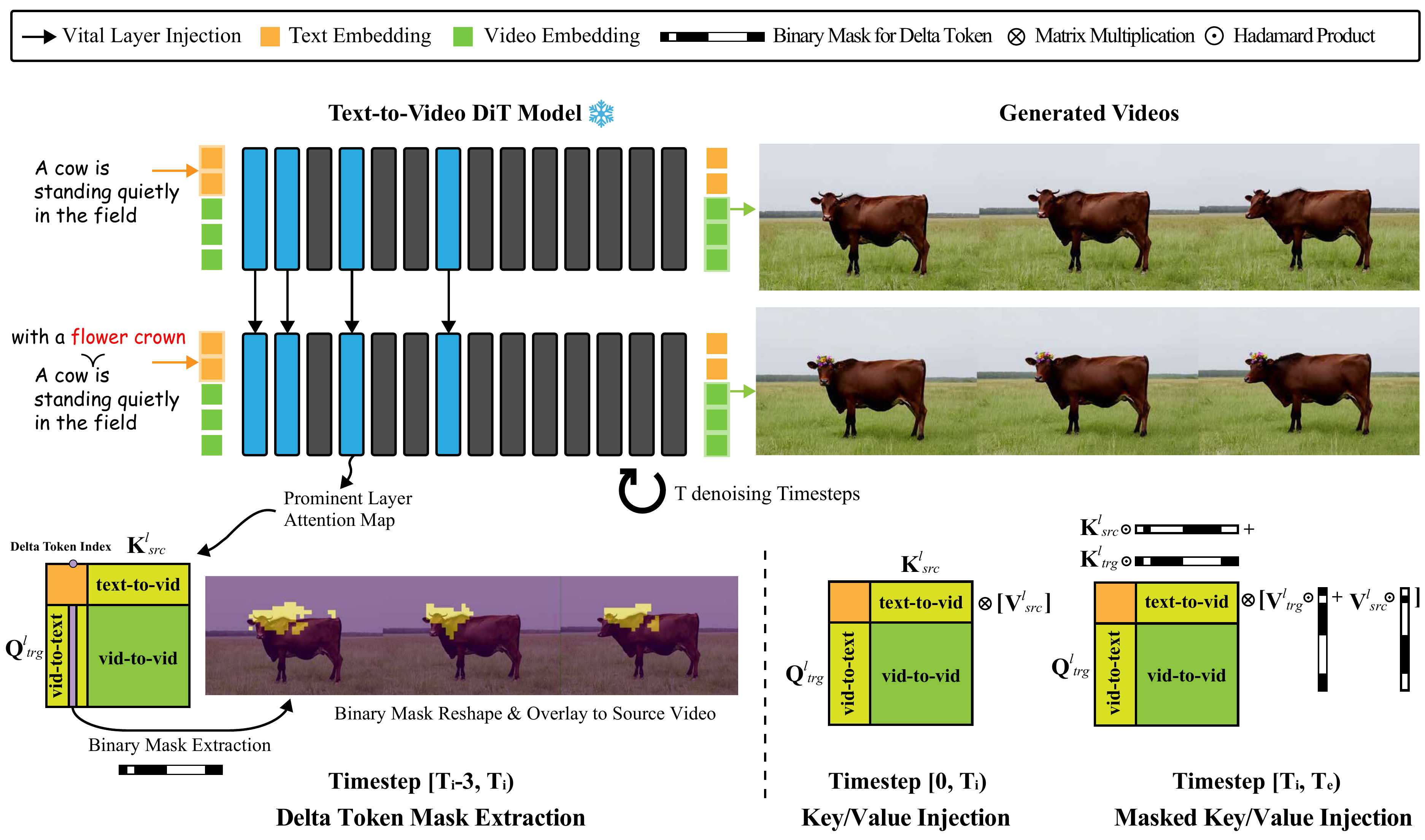}
    \caption{Overview of our object addition framework. Up to the denoising timestep $T_i$, standard key/value injection is applied to vital layers. From $T_i$ to $T_e$, masked key/value injection is performed. The mask is obtained by averaging and then binarizing the attention maps of the prominent layer over the last three timesteps during the standard injection phase.}
    \label{fig:main_method}
    % \includegraphics[width=0.9\linewidth]{figures/prominent_layer.pdf}
    % % \caption{\textbf{Layer Prominence Visualization}. (a) The top-left image is from the source video, and the others show results when RoPE is ablated at each layer. Layer 11, which mainly affects the foreground, has the highest prominence score. (b) Visualization of where the target video attend the most at each denoising timestep, when the key and value of vital layers are injected without mask. Red represents target prompt, yellow represents delta token, and blue represents source video. Around timestep 8, we observed that the target video has finished structuring the video contents and start to attend to delta token (\textcolor{darkyellow}{Yellow}) at proper location. Only middle frame is visualized for the simplicity.}  % 이미지 캡션
    % \caption{\textbf{Layer Prominence Visualization}. (a) The top-left image is from the source video, and the others show results when RoPE is ablated at each layer. Layer 11, which mainly affects the foreground, has the highest prominence score. (b) Visualization of where the target video attends the most at each denoising timestep. \textcolor{red}{Red}: target prompt, \textcolor{yellow}{Yellow}: delta token, \textcolor{blue}{Blue}: source video. Around timestep 8, the target video starts attending to the delta token at the correct location. Only the middle frame is shown for clarity.}
    % \label{fig:prominent_layer}
\end{figure}
%% 마스크를 추출할 layer를 판단하기 위해 layer prominence를 제안
%% fg/bg psnr측정
%% prominence 정의
% 특정 layer가 foreground생성에 영향을 미친다
To identify layers responsible for shaping the foreground, 
we reuse videos generated during the RoPE vitality analysis with a set $S$ of $N_p$ diverse text prompts.
We extract the foreground region of each frame by locating the foreground referent in the prompt (e.g., “lion” in “A lion slowly walking across the savannah during a golden hour sunset”), using Grounded SAM2~\cite{ren2024grounded}.
% , which integrates Grounding DINO~\cite{ren2024grounded} with SAM 2~\cite{ravi2024sam2segmentimages}.
We then compute the PSNR between the sampled video and a video generated by dropping RoPE at a specific layer, averaging the results over spatial and temporal dimensions.
Formally, let \(\overline{PSNR}_{fg}(l)\) and \(\overline{PSNR}_{bg}(l)\) denote the average PSNR values over the foreground and background regions, respectively, for layer \(l\). 
Then, we define the normalized similarity scores for the foreground and background at layer \( l \) as follows:

\begin{equation}
\begin{aligned}
S_{fg}(l) &= 1 - 10^{ - \left( \overline{PSNR}_{fg}(l) - \overline{PSNR}_{min} \right) \left( \frac{\overline{PSNR}_{max}}{C} \right) }, \\
S_{bg}(l) &= 1 - 10^{ - \left( \overline{PSNR}_{bg}(l) - \overline{PSNR}_{min} \right) \left( \frac{\overline{PSNR}_{max}}{C} \right) }.
\end{aligned}
\end{equation}
% $S_{min} = \min \left( \min_{\forall l} \overline{PSNR}_{FG}(l), \min_{\forall l} \overline{PSNR}_{BG}(l) \right)$
% $S_{max} = \max \left( \max_{\forall l} \overline{PSNR}_{FG}(l), \max_{\forall l} \overline{PSNR}_{BG}(l) \right).$
Here, \( \overline{PSNR}_{min} \) and \( \overline{PSNR}_{max} \) represent the minimum and maximum  \(\overline{PSNR}\) values across all layers and regions.
$S_{fg}(l)$ and $S_{bg}(l)$ express similarity using normalized PSNR,
mapping the minimum PSNR to 0 and PSNR values exceeding 40 close to 1. 
$C$ is a hyperparameter set to 400.

Per layer similarity $S_{fg}(l)$ and $S_{bg}(l)$ are
shown in \Cref{fig:layer_analysis}(c) as blue and orange curves.
Since our objective is to identify layers that minimally affect the background while significantly impacting the foreground, 
% To determine the layer that has a high impact only on the foreground region, 
we define Layer Prominence as follows:
\begin{equation}
    P(l) = S_{bg}(l)(1 - S_{fg}(l))
    \label{eq:prominence}
\end{equation},
where $P(l)$ denotes the prominence of layer $l$. 
Layer prominence is visualized as a green curve in ~\Cref{fig:layer_analysis} (c).
Notably, the prominence value is high only when the background similarity $S_{BG}$ is large while the foreground similarity $S_{fg}$ is small.
We refer to layers with high prominence values as \textit{prominent layers}, and those with low values as non-prominent layers.

To better understand their roles, we further analyzed and visualized both prominent and non-prominent layers.
As shown in \Cref{fig:layer_analysis} (c) and \Cref{fig:prominent_layer} (a), layers exhibiting high similarity in both foreground and background regions, or conversely, low similarity in both, tend to have low prominence.
In contrast, when the RoPE is dropped, layer 11 maintain background similarity reasonably well while significantly altering the foreground, resulting in a high prominence value. 
Based on this observation, we identify layer 11 as the most prominent layer responsible for foreground generation and utilize it for mask extraction.

% 11번 레이어가 제일 높은 prominence
% 시각적으로도 그러함.
% 마스크는 다음과 같은 수식에 의해 추출
%% attention scaling
%% gaussian filtering
%% thresholding
% 
% 우리는 논문 전반에 걸쳐 11번 레이어에서 추출한 마스크 외 영역에 대해 동일한 파라미터를 가지고 key value injection을 수행함.
% \begin{wrapfigure}{r}{0.4\textwidth}  % 'r'은 오른쪽 정렬, 0.5\textwidth는 페이지 가로 길이의 절반
%     % \captionsetup{belowskip=10pt} % 캡션 아래 공간 확보, 텍스트가 침범하지 못하도록    
%     \centering
%     \includegraphics[width=\linewidth]{figures/layer_prominence.pdf}
%     \caption{\textbf{Layer Prominence Visualization}. The top-left image is sampled from the source video, while the remaining images show results when RoPE is ablated at each individual layer during inference. Among them, layer 11, which primarily influences the foreground, exhibits a higher prominence score.}
%     \label{fig:prominence}
%     % \end{minipage}
%     % \includegraphics[width=0.45\textwidth]{example-image}  % 이미지 크기 설정
%     \caption{Example image}  % 이미지 캡션
% \end{wrapfigure}
\noindent \textbf{Masked Key/Value Injection}
~\Cref{fig:main_method} illustrates the overall framework of our method.
For object addition, we inject the key and value tensors of vital layers from the DiT video generation model of the source video into those of the target video,
formulated as $Attn_{t}(Q^{l}_{trg}, K^{l}_{src}, V^{l}_{src})$,
% formulated as:
% \begin{equation}
%     Attn_{t}(Q^{l}_{trg}, K^{l}_{src}, V^{l}_{src})
% \end{equation},
where $Attn_{t}(Q, K, V)$ denotes the self-attention operation on visual features at denoising timestep \(t\), 
with \(Q^{l}_{trg}\) as the query from the target video at layer \(l\),  
and \(K^{l}_{src}\), \(V^{l}_{src}\) as the key and value from the source video at the same layer.
We perform key/value injection from timestep 0 up to \(T_i - 1\).  
However, as discussed in ~\Cref{sec:vitality}, simply injecting keys and values is insufficient for video editing. 
So we performed mask extraction from the prominent layer during the time step of  $T_i-3$, $T_i-2$, and $T_i-1$.
% The attention map of the prominent layer from the last three timesteps before \(T_i\) is used for mask extraction.  
To extract the mask, we utilize \textit{delta tokens}, which correspond to the tokens that differ between the source and target prompts. 
For instance, in ~\Cref{fig:main_method}, the tokens corresponding to “flower crown” are considered delta tokens. 
The mask for the \textit{delta tokens} is obtained by averaging these attention maps and thresholding them using the value \(t_{mask}\).
Further details on the mask extraction process can be found in~\Cref{appsec:objadd}, specifically in~\Cref{alg:objadd}.

The choice of timesteps for mask extraction is determined empirically. 
As shown in ~\Cref{fig:prominent_layer}(b), as the denoising timestep increases,
the target video gradually attends only to the source video, ignoring the target prompt.  
Meanwhile, the target video shows negligible changes after timestep 25.  
We define the \textit{prominent timestep} as 8, where there is room for editing and where the delta tokens attend to the appropriate locations.  
% Therefore, we extract the mask using the attention maps of the prominent layer at the prominent timestep and its adjacent timesteps.
Therefore, we extract the mask using the prominent layer’s attention maps around the prominent timestep.
The extracted mask \(M_{obj}\) is used for masked key/value injection as $Attn_{t}(Q^{l}_{trg}, K^{l}_{mix}, V^{l}_{mix})$,
% The extracted mask \(M_{obj}\) is used for masked key/value injection:
% \begin{equation}
%     Attn_{t}(Q^{l}_{trg}, K^{l}_{mix}, V^{l}_{mix})
% \end{equation},
where $K^{l}_{mix} = M_{obj} K^{l}_{trg} + (1 - M_{obj}) K^{l}_{src}$, $V^{l}_{mix} = M_{obj} V^{l}_{trg} + (1 - M_{obj}) V^{l}_{src}$.
% $K^{l}_{mix}=M_{obj}~\times K^{l}_{trg} + (1-M_{obj})~\times K^{l}_{src}$ and $V^{l}_{mix}=M_{obj}~\times V^{l}_{trg} + (1-M_{obj})~\times V^{l}_{src}$.
We confirmed that injecting the source video only into the non-masked region prevents the target video from replicating the source video.

\subsection{Video Editing for Non-Rigid Deformation}
\label{sec:non_rigid}
For non-rigid editing, we need to inject appearance information from the source video, 
while allowing room for the object deformation.
To this end, we inject key/value features into non-vital, position-agnostic layers, effectively decoupling spatial alignment from semantic transfer.
By injecting the key and value tensors of non-vital layers from the source video generation model during timesteps \([0, T_e)\) without applying a mask, we confirmed that non-rigid editing is achievable.
% We confirmed that non-rigid editing is achievable by injecting the key and value tensors from non-vital layers of the source video generation model. This injection is performed during the timesteps \([0, T_e)\) without applying a mask.
% This simple yet effective design choice allows the model to flexibly transfer motion characteristics, demonstrating strong editing capability with minimal intervention to the vital layer.
This simple yet effective design choice enables the model to adapt to the motion semantics conveyed by the prompt, demonstrating strong editing capability with minimal intervention to the vital layer.

\section{Experiments}
% 선택한 hyper param들 언급.
% 평가 대상과 방법 언급.

% \noindent \textbf{Mast Extraction Timestep}
% \begin{figure}[t]
%     \centering
%     \includegraphics[width=\linewidth]{figures/mask_extraction_timestep.pdf}
%     \caption{Example image}  % 이미지 캡션
% \end{figure}

\subsection{Experimental Details}

\begin{wrapfigure}{r}{0.45\textwidth}  % 'r'은 오른쪽 정렬, 0.5\textwidth는 페이지 가로 길이의 절반
    \centering
    \includegraphics[width=\linewidth]{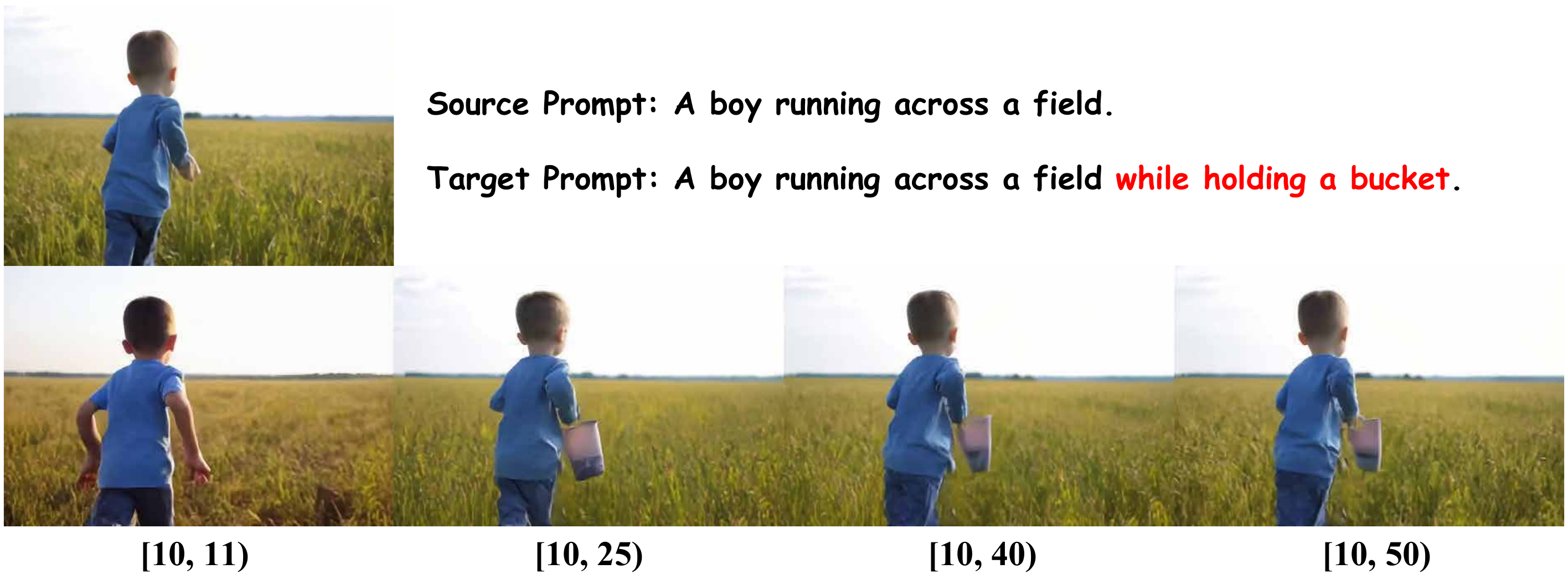}    
    \caption{Denoising timesteps ablation where the masked injection is applied. First Row: Sampled frame from source video. Second row: Generated video frame when the masked injection is applied in range$[\cdot, \cdot)$}  % 이미지 캡션
    \label{fig:timestep_ablation}
\end{wrapfigure}

% Perform key value injection without mask in $[0, T_{n})$ range.
% Extract mask from $[T_{n-3}, T_{n})$ timesteps.\\
We use CogVideoX~\cite{yang2024cogvideox} as our base text-to-video generation model.
For object addition task, we perform key/value masked injection during $[T_{i}, T_{e})$ denoising timestep range, where the $T_{e}$, which is 25 in our case, is different from the total denoising step $T$.
It is because, as we can see from ~\Cref{fig:prominent_layer} (b), denoising steps after 30 barely affect the final generation results. 
Secondly, as shown in ~\Cref{fig:timestep_ablation}, early-only injection leads to semantic drift, while injection beyond timestep 40 fails to achieve natural object fusion.
Masked injection from timestep [10, 25) is the sweetspot that we found and achieves non-masked region preservation and natural object fusion.
The number of text prompts used for layer analysis, \(N_p\), is 40.
We set the mask extraction threshold $t_{mask}$ to $0.8$ and 
Vital layers we used are [0, 1, 10, 11, 12, 14, 15, 17, 19, 23], and non-vital layers that we used are [16, 24, 25, 27, 28, 29, 30, 31, 32, 33, 34, 35, 36, 37, 38, 39, 40, 41].
All of the aforementioned values are applied consistently across all experimental results presented in this paper.

\subsection{Quantitative Results}
\label{sec:quan}
\begin{table*}[t]
\captionsetup{type=table}
\caption{Comparison of methods on the object addition and non-rigid editing. We multiply \textbf{CLIP$_{dir}$}, \textbf{CLIP$_{img}$}, Temporal Flickering (\textbf{T.F.}), Motion Smoothness (\textbf{M.S.}), Subject Consistency (\textbf{S.C.}), and Background Consistency (\textbf{B.C.}) to compute the overall evaluation score. Best results are shown in \textbf{bold}, and second-best results are \underline{underlined}.}

\centering
\resizebox{\textwidth}{!}{%
\begin{tabular}{ll|cc|c|cccc|c}
\toprule
\textbf{Task} & \textbf{Method} & \textbf{CLIP$_{dir}$~$\uparrow$} & \textbf{CLIP$_{img}$~$\uparrow$} & \textbf{CLIP$_{all}$~$\uparrow$} & \textbf{T.F.~$\uparrow$} & \textbf{M.S.~$\uparrow$} & \textbf{S.C.~$\uparrow$} & \textbf{B.C.~$\uparrow$} & \textbf{Overall~$\uparrow$} \\
\midrule

\multirow{7}{*}{\textbf{Object Addition}} 
& BIVDiff~\cite{shi2024bivdiff} & 0.0940 & 0.7734 & 0.2425 & 0.8980 & 0.9120 & 0.8060 & 0.9330 & 0.1494 \\
& RAVE~\cite{kara2024rave}& 0.0456 & 0.8407 & 0.2406 & 0.9560 & 0.9700 & \underline{0.9400} & \textbf{0.9710} & 0.2036 \\
& CogInv~\cite{yang2024cogvideox} & 0.0262 & \textbf{0.9421} & 0.2491 & \textbf{0.9790} &\textbf{0.9870} & \textbf{0.9420} & 0.9510 & 0.2156 \\
& VidToMe~\cite{li2024vidtome}& 0.1130 & 0.8392 & 0.2559 & 0.9600 & 0.9720 & 0.9190 & \underline{0.9540} & 0.2093 \\
& CogV2V~\cite{yang2024cogvideox} & \underline{0.1167} & 0.9042 & \underline{0.2658} & \underline{0.9720} & \underline{0.9850} & 0.9390 & 0.9530  & \underline{0.2277} \\

\cmidrule(lr){2-10}
& \modelname{} (\text{\scriptsize w/ N.P. Mask}) & 0.0534 & \underline{0.9361} & 0.2549 & 0.9664 & 0.9808 & 0.9336 & 0.9507 & 0.2145\\
& \textbf{\modelname} & \textbf{0.1258} & {0.9294} & \textbf{0.2715} & 0.9660 & 0.9810 & 0.9340 & 0.9450  & \textbf{0.2286} \\
\midrule

\multirow{7}{*}{\textbf{Non-Rigid}} 
& BIVDiff~\cite{shi2024bivdiff} & 0.0213 & 0.8282 & 0.2334 & 0.9150 & 0.9250 & 0.8700 & 0.9550 &  0.1641 \\

& RAVE~\cite{kara2024rave}& 0.0007 & 0.8759 & 0.2347 & 0.9620 & 0.9730 & \underline{0.9670} & \textbf{0.9830} & 0.2088 \\

& CogInv~\cite{yang2024cogvideox} & -0.0064 & \underline{0.9561} & 0.2429 & \textbf{0.9830} & \textbf{0.9900} & \underline{0.9670} & 0.9620 & 0.2199 \\

& VidToMe~\cite{li2024vidtome} & \underline{0.0585} & 0.8868 & 0.2496 & 0.9720 & 0.9810 & 0.9570 & 0.9670 & 0.2203 \\

& CogV2V~\cite{yang2024cogvideox}& 0.0284 & 0.9280 & 0.2478 & \underline{0.9790} & \underline{0.9880} & 0.9660 & 0.9670 & 0.2239 \\

\cmidrule(lr){2-10}
& \modelname{} (\text{\scriptsize w/ V.L.}) & 0.0113 & \textbf{0.9681} & \underline{0.2488} & 0.9758 & 0.9865 & \textbf{0.9688} & \underline{0.9694} & \underline{0.2249} \\
& \textbf{\modelname} & \textbf{0.0821} & 0.9015 & \textbf{0.2572} & 0.9750 & 0.9850 & 0.9540 & 0.9570 & \textbf{0.2255} \\

\bottomrule
\end{tabular}%
}

\label{tab:quant}
\end{table*}

We compare our results with state-of-the-art video editing methods, where CogV2V and CogInv are CogVideoX~\cite{yang2024cogvideox} supported SDEdit-based and inversion-based video editing methods, respectively. VidToMe~\cite{li2024vidtome}, BIVDiff~\cite{shi2024bivdiff}, and RAVE~\cite{kara2024rave} are state-of-the-art training-free video editing methods.
To define test set, we use ChatGPT~\cite{openai2023chatgpt} to generate text prompt pairs that are suitable for object addition and non-rigid editing tasks. 30 prompt pairs are generated for each task.
Using the same set of prompt pairs, we conduct comparison with baseline methods. 
For the measurement, we use the following metrics.
\textbf{CLIP$_{dir}$}~\cite{gal2022stylegan} calculates the CLIP similarity between direction of text change and direction of averaged video frame change, which best suits for our editing task evaluation, \textbf{CLIP$_{img}$} measures average similarity between source video frames and target video frames. Note that \textbf{CLIP$_{img}$} metric can have high value when the edited video replicates the source video. To consider whether the edited video is similar to the source video while faithfully adapt target text prompt at once, we multiply \textbf{CLIP$_{dir}$} and \textbf{CLIP$_{img}$}, producing \textbf{CLIP$_{all}$}. \textbf{CLIP$_{all}$} metric shows that our method best suits our purpose. 
Meanwhile, \textbf{CLIP} based metrics are the average of framewise metrics, and lack of temporal consistency. 
To consider temporal consistency, we adopt VBench~\cite{huang2024vbench} video evaluation metrics, which are Temporal Flickering (\textbf{T.F.}), Motion Smoothness (\textbf{M.S.}), Subject Consistency (\textbf{S.C.}), and Background Consistency (\textbf{B.C.}).
However, these metrics can also have high value when the edited video replicates the source video.
To consider prompt direction alignment, source video similarity, and video quality all at once, we multiply them and consider it as integrated metric. 
~\Cref{tab:quant} shows that our method outperforms all baseline method in overall metrics and \textbf{CLIP$_{all}$} metric regardless of the task.

\noindent \textbf{Ablation Study}
% We also conducted ablation study by extracting mask from non-prominent(N.P.) layer, which is 33th layer, and denoting the result in \modelname (w/ N.P) for object addition. As we can see from the values in ~\Cref{tab:quant}, we can conjecture that extracting the mask from the prominent layer was crucial for object addition.
We conduct additional ablation experiments to validate our approach. 
For object addition, we extract masks from a non-prominent (N.P.) layer (specifically, the 33rd layer) and refer to this configuration as \modelname (w/ N.P.).
As shown in \Cref{tab:quant}, the significant performance drop confirms that extracting masks from prominent layers is crucial for successful object addition.
% For non-rigid editing, ablation experiments are conducted by injecting the key/value of vital layers instead of non-vital layers, denoting it as \modelname (w/ V.L.). We found that this can only replicate the source video as it is, losing deformation editing capability. However, since the metrics except \textbf{CLIP$_{dir}$} tend to give a high score to the result that replicates the source video, it still show high scores in  \textbf{CLIP$_{img}$} metric and VBench metrics, though small \textbf{CLIP$_{dir}$} metric stop it from having high overall score.
For non-rigid editing, we perform ablation by injecting key/value features of vital layers instead of non-vital layers, denoted as \modelname (w/ V.L.). This configuration merely replicates the source video without achieving the desired deformation capabilities. Interestingly, while this approach scores high on \textbf{CLIP$_{img}$} and VBench metrics (which favor source video preservation), its substantially lower \textbf{CLIP$_{dir}$} score—which measures directional alignment with editing instructions—prevents it from achieving a high overall performance. This confirms that our layer selection strategy is critical for enabling meaningful non-rigid transformations while maintaining video coherence.

\begin{wrapfigure}{r}{0.48\textwidth}
\captionsetup{type=table}
\caption{User Study Results. Values represent user preference percentages (\%).}
\label{tab:user}
\scriptsize % 더 작은 글자 크기
\renewcommand{\arraystretch}{1.0} % 행간 압축
\begin{tabularx}{0.47\textwidth}{
    >{\raggedright\arraybackslash}p{1.1cm} % Method 열 고정폭 + 줄바꿈 허용
    *{3}{>{\centering\arraybackslash}X}|*{3}{>{\centering\arraybackslash}X}
}
\toprule
& \multicolumn{3}{c|}{\textbf{Object Addition}} & \multicolumn{3}{c}{\textbf{Non-Rigid}} \\
Methods & Q1 & Q2 & Q3 & Q1 & Q2 & Q3 \\
\midrule
BivDiff~\cite{shi2024bivdiff}  & 0.2 & 0.2 & 1.0 & 0.2 & 0.0 & 0.0 \\
RAVE~\cite{kara2024rave}     & 0.8 & 1.4 & 3.9 & 0.2 & 0.4 & 1.9 \\
VidToMe~\cite{li2024vidtome}  & 2.7 & 2.2 & 3.9 & 1.0 & 0.8 & 1.1 \\
CogInv~\cite{yang2024cogvideox}   & 1.4 & 1.4 & 3.9 & 2.9 & 3.9 & 24.2 \\
CogV2V~\cite{yang2024cogvideox}      & 19.6 & 15.3 & 11.2 & 4.7 & 7.3 & 21.4 \\
\textbf{\modelname}     & \textbf{75.3} & \textbf{79.6} & \textbf{78.6} & \textbf{91.0} & \textbf{87.6} & \textbf{51.5} \\
\bottomrule
\end{tabularx}
\end{wrapfigure}

\noindent \textbf{User Study}
We conduct a user study comparing six different methods. 
For each of the two tasks—object addition and non-rigid editing—we present 15 distinct examples. 
Participants answer three preference questions for each example by selecting the method they prefer most. 
The study involves 34 participants evaluating video quality based on three predefined criteria. 
As summarized in ~\Cref{tab:user}, our method, \modelname{}, receives significantly higher user preference scores in both object addition (75.3-79.6\%) and non-rigid editing tasks (51.5-91.0\%). The full set of survey questions is provided in the supplementary material.
% For each of the two tasks, object addition and non-rigid editing, 15 examples is presented. For each example, participants answered three preference questions by selecting the method they preferred the most. A total of 34 participants took part in the study. As summarized in ~\Cref{tab:user}, our method, \modelname{}, received the highest user preference scores in both the object addition and the non-rigid editing tasks. The full set of survey questions is provided in the supplementary material.

% User Study
\subsection{Qualitative Results}
\noindent \textbf{Experiments on Real-World Video}

\begin{wrapfigure}{r}{0.45\textwidth}  % 'r'은 오른쪽 정렬, 0.5\textwidth는 페이지 가로 길이의 절반   
    \vspace{-5em}
    \centering
    \includegraphics[width=\linewidth]{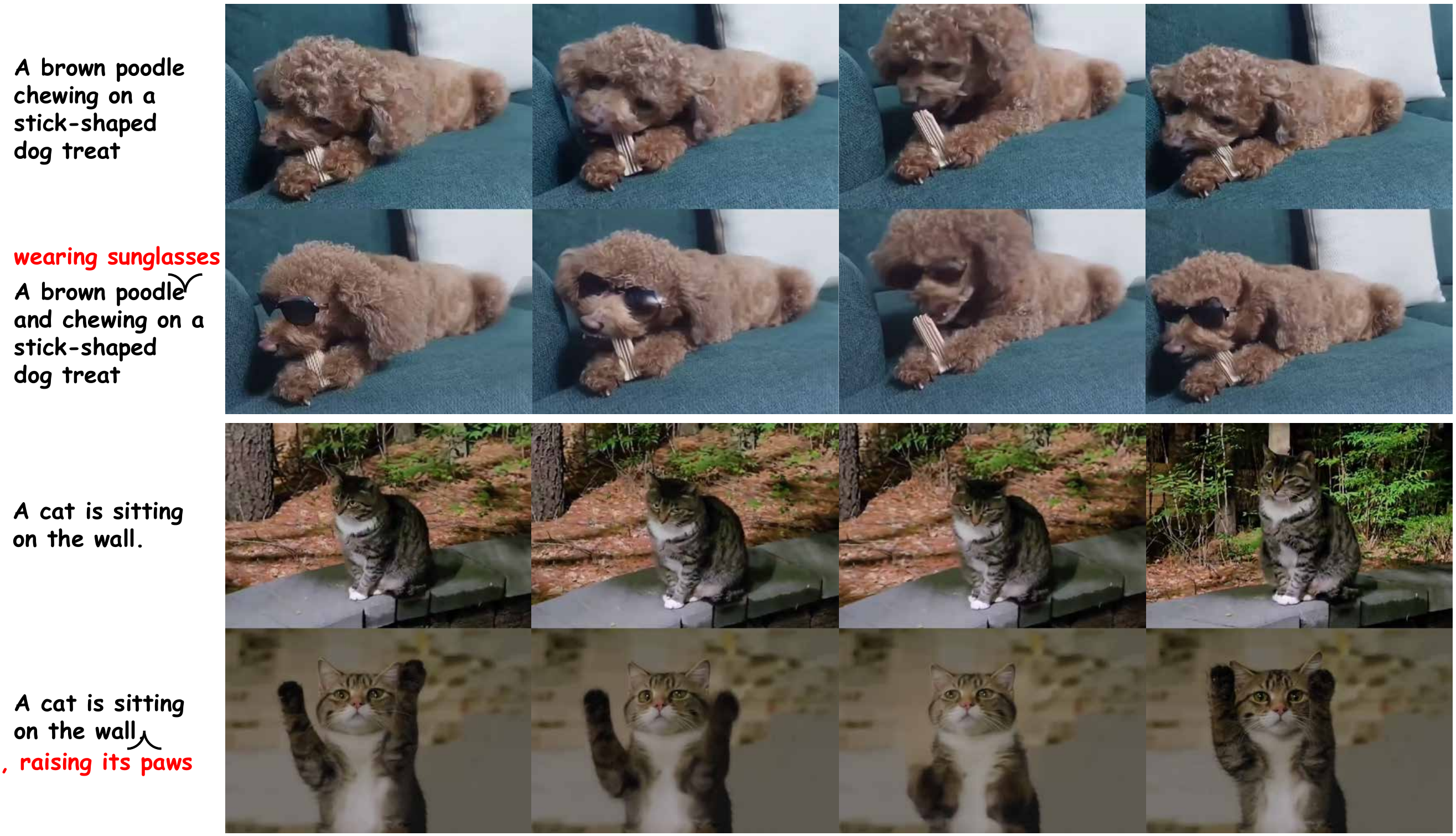}    
    \caption{Real video editing results. The top shows object addition, and the bottom shows non-rigid editing.}  % 이미지 캡션
    \label{fig:real}    
\end{wrapfigure}
% Our method is applicable to real world video editing. 
% Given real-world video, we apply DDIM inversion~\cite{song2020denoising} to the video, to find reliable latent noise trajectory of the given video. Using this inversion result, we apply same
Our method is applicable to real-world video editing.
Given real-world video, we apply DDIM inversion~\cite{song2020denoising} to the video, 
% to find reliable latent noise trajectory of the given video. 
to find initial noise of the given video. 
Starting from this inversion result, we apply the same~\modelname{} approach.
As shown in ~\Cref{fig:real}, our method is able to perform object addition and non-rigid editing using real-world video.
We use casually captured video using a cell phone for these experimental results.
Note that we use the same hyperparameters throughout the paper, including the real-world data results.
\begin{figure}[t]
\includegraphics[width=0.95\linewidth]{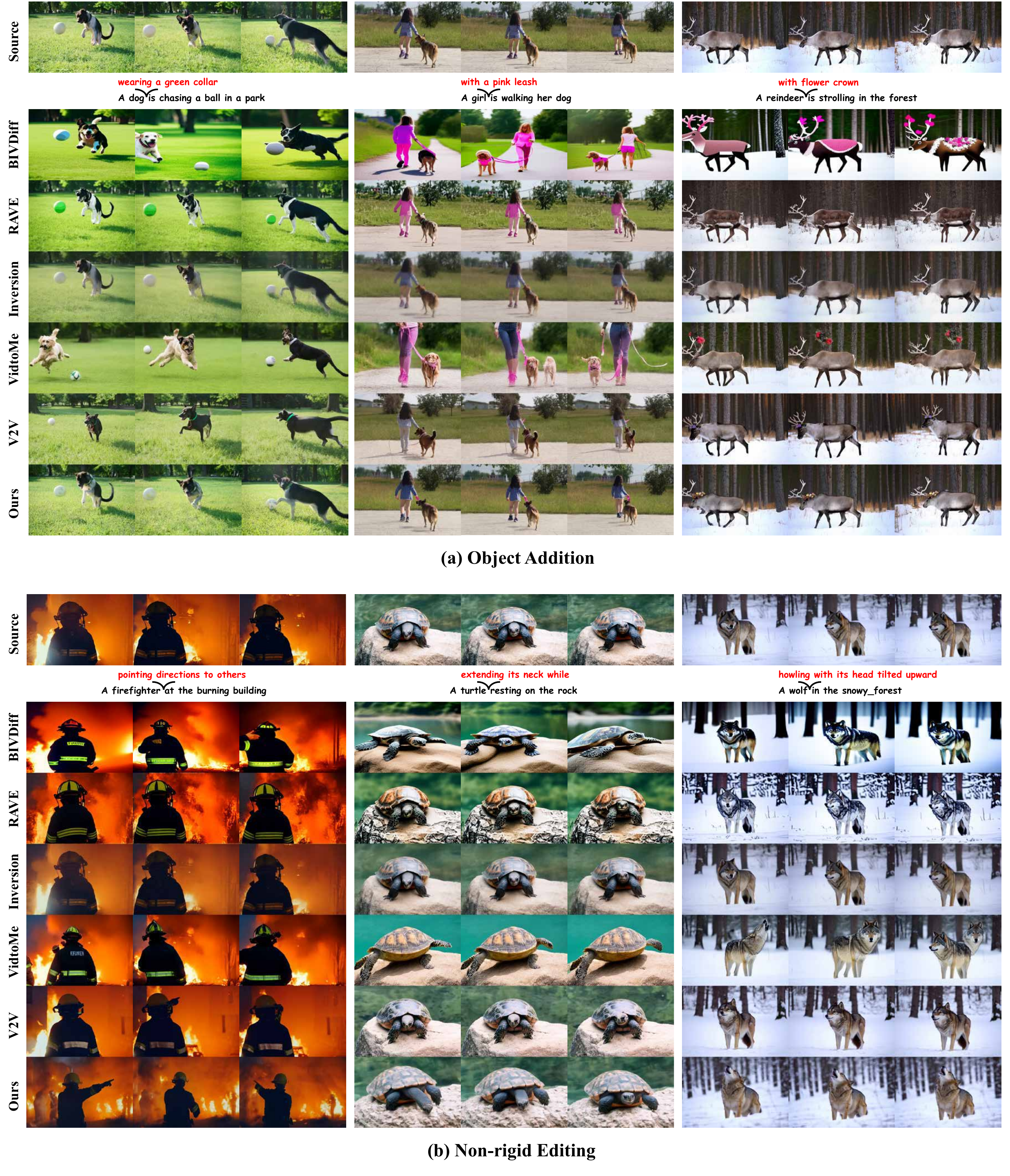}
\caption{Visual comparison between our method and baselines on (a) object addition and (b) non-rigid editing. }
\label{fig:qual_all}
\end{figure}

\noindent\textbf{Visual Comparison}
\Cref{fig:qual_all} shows the qualitative comparisons with baseline methods. 
We observe that our method not only synthesizes newly added objects effectively in the object addition task, but also preserves the surrounding regions with minimal distortion.
Furthermore, qualitative results show that our method effectively reflects the newly introduced non-rigid prompts, while preserving appearance information from the source video without simply copying it. This demonstrates our model's ability to transfer content in a controllable and faithful manner.
\section{Conclusion}
In this paper, we present an in-depth analysis of the layer-wise characteristics of DiT-based video models. Building on these observations, we propose a training-free method capable of performing object addition and non-rigid video editing. We hope that our findings provide a useful perspective on DiT-based video editing and contribute to future developments in both training-free and learning-based approaches.

\newpage
\section*{NeurIPS Paper Checklist}

\begin{enumerate}

\item {\bf Claims}
    \item[] Question: Do the main claims made in the abstract and introduction accurately reflect the paper's contributions and scope?
    \item[] Answer: \answerYes{} % Replace by \answerYes{}, \answerNo{}, or \answerNA{}.
    \item[] Justification: Yes. The abstract and introduction clearly state the main contributions, including the training-free text-guided video editing framework, analysis of DiT model layers, and empirical performance improvements. These claims are consistent with the methods and results presented in the paper (see Sections 3–5).%\justificationTODO{}

\item {\bf Limitations}
    \item[] Question: Does the paper discuss the limitations of the work performed by the authors?
    \item[] Answer: \answerYes{} % Replace by \answerYes{}, \answerNo{}, or \answerNA{}.
    \item[] Justification: We include a dedicated “Limitations” section in the supplemental material. %, where we discuss key assumptions, generalization boundaries, and computational constraints.%\justificationTOD{}

\item {\bf Theory assumptions and proofs}
    \item[] Question: For each theoretical result, does the paper provide the full set of assumptions and a complete (and correct) proof?
    \item[] Answer: \answerYes{}{} % Replace by \answerYes{}, \answerNo{}, or \answerNA{}.
    \item[] Justification: We show that proper layer selection based on vitality and prominence improves video editing performance, supported by quantitative metrics.

    \item {\bf Experimental result reproducibility}
    \item[] Question: Does the paper fully disclose all the information needed to reproduce the main experimental results of the paper to the extent that it affects the main claims and/or conclusions of the paper (regardless of whether the code and data are provided or not)?
    \item[] Answer: \answerYes{} % Replace by \answerYes{}, \answerNo{}, or \answerNA{}.
    \item[] Justification: We provide a complete list of all hyperparameters in the paper and will release the code upon acceptance to enable full reproducibility of our main experimental results.

\item {\bf Open access to data and code}
    \item[] Question: Does the paper provide open access to the data and code, with sufficient instructions to faithfully reproduce the main experimental results, as described in supplemental material?
    \item[] Answer: \answerYes{} % Replace by \answerYes{}, \answerNo{}, or \answerNA{}.
    \item[] Justification: We will publicly release the full codebase upon acceptance, with detailed installation and usage instructions provided in the supplemental material.

\item {\bf Experimental setting/details}
    \item[] Question: Does the paper specify all the training and test details (e.g., data splits, hyperparameters, how they were chosen, type of optimizer, etc.) necessary to understand the results?
    \item[] Answer: \answerYes{} % Replace by \answerYes{}, \answerNo{}, or \answerNA{}.
    \item[] Justification: We specify all training and test details, including using CogVideoX as the base model, the denoising timestep range ([10, 25)), mask threshold ($t_{mask}=0.8$), and the exact lists of vital and non-vital layers; see Sec. 5.% \justificationTODO{}

\item {\bf Experiment statistical significance}
    \item[] Question: Does the paper report error bars suitably and correctly defined or other appropriate information about the statistical significance of the experiments?
    \item[] Answer: \answerNo{} % Replace by \answerYes{}, \answerNo{}, or \answerNA{}.
    \item[] Justification: Measuring error bars is uncommon in video generation research due to the substantial memory demands and lengthy computation times, which hinder the collection of enough data to reliably estimate error bars given the nature of these models.

\item {\bf Experiments compute resources}
    \item[] Question: For each experiment, does the paper provide sufficient information on the computer resources (type of compute workers, memory, time of execution) needed to reproduce the experiments?
    \item[] Answer: \answerYes{} % Replace by \answerYes{}, \answerNo{}, or \answerNA{}.
    \item[] Justification: We include a dedicated "computational cost" in the supplemental metarial.%\justificationTODO{}
    
\item {\bf Code of ethics}
    \item[] Question: Does the research conducted in the paper conform, in every respect, with the NeurIPS Code of Ethics \url{https://neurips.cc/public/EthicsGuidelines}?
    \item[] Answer: \answerYes{} % Replace by \answerYes{}, \answerNo{}, or \answerNA{}.
    \item[] Justification: We target training-free video editing and therefore did not use any external training data; all evaluation videos were ethically generated by our video generation model, and any real-world videos employed are from copyright-compliant sources. %\justificationTODO{}

\item {\bf Broader impacts}
    \item[] Question: Does the paper discuss both potential positive societal impacts and negative societal impacts of the work performed?
    \item[] Answer: \answerYes{} % Replace by \answerYes{}, \answerNo{}, or \answerNA{}.
    \item[] Justification: Our training-free method democratizes video editing—enabling broader access to creative tools and lowering production barriers—while avoiding any additional fine-tuning or training significantly reduces computational energy use and associated environmental impact.%\justificationTODO{}
    
\item {\bf Safeguards}
    \item[] Question: Does the paper describe safeguards that have been put in place for responsible release of data or models that have a high risk for misuse (e.g., pretrained language models, image generators, or scraped datasets)?
    \item[] Answer: \answerNA{} % Replace by \answerYes{}, \answerNo{}, or \answerNA{}.
    \item[] Justification: Our method does not involve data release, and since it is a training-free approach, this question is not applicable.

\item {\bf Licenses for existing assets}
    \item[] Question: Are the creators or original owners of assets (e.g., code, data, models), used in the paper, properly credited and are the license and terms of use explicitly mentioned and properly respected?
    \item[] Answer: \answerYes{} % Replace by \answerYes{}, \answerNo{}, or \answerNA{}.
    \item[] Justification: We use the pretrained CogVideoX-5B~\cite{yang2024cogvideox} model as a baseline, which is released by the CogVideoX Model Team under a non-commercial academic research license. The license explicitly allows free academic use and prohibits military or illegal applications.%\justificationTODO{}

\item {\bf New assets}
    \item[] Question: Are new assets introduced in the paper well documented and is the documentation provided alongside the assets?
    \item[] Answer: \answerYes{} % Replace by \answerYes{}, \answerNo{}, or \answerNA{}.
    \item[] Justification: We will release all new assets, including code and documentation, upon acceptance. The release will include structured details on model usage, limitations, and license terms, and will be anonymized appropriately during the submission phase.%\justificationTODO{}

\item {\bf Crowdsourcing and research with human subjects}
    \item[] Question: For crowdsourcing experiments and research with human subjects, does the paper include the full text of instructions given to participants and screenshots, if applicable, as well as details about compensation (if any)? 
    \item[] Answer: \answerNA{} % Replace by \answerYes{}, \answerNo{}, or \answerNA{}.
    \item[] Justification: Our work does not involve any crowdsourcing or research with human subjects. %\justificationTODO{}

\item {\bf Institutional review board (IRB) approvals or equivalent for research with human subjects}
    \item[] Question: Does the paper describe potential risks incurred by study participants, whether such risks were disclosed to the subjects, and whether Institutional Review Board (IRB) approvals (or an equivalent approval/review based on the requirements of your country or institution) were obtained?
    \item[] Answer: \answerNA{} % Replace by \answerYes{}, \answerNo{}, or \answerNA{}.
    \item[] Justification: Our work does not involve any crowdsourcing or research with human subjects.%\justificationTODO{}

\item {\bf Declaration of LLM usage}
    \item[] Question: Does the paper describe the usage of LLMs if it is an important, original, or non-standard component of the core methods in this research? Note that if the LLM is used only for writing, editing, or formatting purposes and does not impact the core methodology, scientific rigorousness, or originality of the research, declaration is not required.
    %this research? 
    \item[] Answer: \answerYes{} % Replace by \answerYes{}, \answerNo{}, or \answerNA{}.
    \item[] Justification: We used ChatGPT solely to generate text prompts for evaluation video generation; it did not play any role in the core methodology, model design, or scientific contributions of the research.%\justificationTODO{}

\end{enumerate}

% emjay added for ref ----------------
\bibliographystyle{unsrtnat}  % or plainnat, abbrvnat, etc.
\bibliography{neurips_2025}
% ------------------------------------
\clearpage
% A Technical Appendices and Supplementary Material
% Technical appendices with additional results, figures, graphs and proofs may be submitted with the paper submission before the full submission deadline (see above), or as a separate PDF in the ZIP file below before the supplementary material deadline. There is no page limit for the technical appendices.
\appendix

\begin{center}
    % \vspace*{1cm}            
    \Large
    \textbf{Supplementary Material}
    % \vfill
\end{center}

\noindent \textbf{Note on Notation Error} \\
We would like to note that there is an inconsistency in the notation used in the main paper. Specifically, the variable $T_i$ was mistakenly written as $T_n$ or $T_m$ in some parts of the paper. These should all consistently refer to $T_i$.  
Also, in ~\Cref{sec:obj_add}, the sentence ``PSNR values exceeding 40 close to 0'' should be ``PSNR values exceeding 40 are mapped to values close to 1'', not 0.  
We apologize for any confusion this may have caused and will ensure that this is corrected in a future revision or the camera-ready version.

\section{Pseudo-Algorithm}

\begin{table}[h]
\centering
\renewcommand{\arraystretch}{1.3}
\begin{tabular}{lp{11cm}}
\toprule
\textbf{Notation} & \textbf{Description} \\
\midrule
\textsc{DM}$(z_t, \mathcal{P}, t, s)$ & Denoising module at step $t$ that outputs denoised latent $z$ and attention map $A$. \\$A_{w^*}$ & The attention map corresponding to the words $w^*$ (also referred to as ``delta tokens'') that specify the object to be added. \\
$vl$ & $vl \in \mathcal{L}$ from the vital layer set $\mathcal{L}.$\\
$nl$ & $nl \in \mathcal{N}$ from the non-vital layer set $\mathcal{N}.$\\
\textsc{WS}$(\cdot)$ & Mask-guided weighted sum for keys/values. For a component $vl$, and a binary mask $M$, it computes:
\[
K^{vl}_{trg} = (1 - M) \odot K^{vl}_{src} + M \odot K^{vl}_{trg}, \quad
V^{vl}_{trg} = (1 - M) \odot V^{vl}_{trg} + M \odot V^{vl}_{src}
\]
Here, $\odot$ denotes element-wise multiplication.\\
\textsc{Normalize}$(\cdot)$ & Normalize input to the $[0, 1]$ range. \\
\textsc{Preprocess}$(\cdot)$ & Preprocesses the attention map. (1) Rescale attention based on the formula $y = \min\left(1, \frac{\log (kx + 1)}{\log(c_k \cdot k + 1)}\right)$. This maps values above a constant $c_k$ to 1 and 0 to 0. 
We use $k=10.0$ and $c_k=0.1$. (2) Apply gaussian blur (kernel size$= 3, \sigma=1.0$). (3) Binarize at threshold $0.8$. \\
\textsc{$[\cdot, \cdot]$} & Batch concatenation. \\
\textsc{$T$} & Total diffusion steps. \\
\textsc{$T_i$} & Diffusion step when standard key/value injection ends (see~\Cref{fig:main_method}). \\
\textsc{$T_e$} & Diffusion step when masked key/value injection ends (see~\Cref{fig:main_method}). \\
% \textsc{len}$(\mathcal{P}^*_{\text{embed}})$ & Length of embedded tokens in target prompt. \\
\textsc{idx}$_{w^*}$ & Indices of tokens corresponding to the words $w^*$ in the target prompt $\mathcal{P}^*$. \\
\textsc{$\mathcal{D}$} & VAE Decoder. \\
\bottomrule
\end{tabular}
% \caption{Notation used in the video editing algorithm.}
\end{table}

\subsection{Object Addition}
\label{appsec:objadd}
\begin{algorithm}[h]
\caption{~\modelname~Video Editing - Object Addition}
\begin{algorithmic}[1]
\State \textbf{Input:} A source prompt $\mathcal{P}$, a target prompt $\mathcal{P}^*$, words $w^*$ in $\mathcal{P}^*$, specifying the object to add, and a random seed $s$.
\State \textbf{Output:} A source video $\mathcal{V}_{\text{src}}$ and an edited video $\mathcal{V}_{\text{trg}}$.
\State $z_0 \sim \mathcal{N}(0, I)$ \Comment{a unit Gaussian random variable with random seed $s$}
\State $z^*_0 \gets z_0$
\State $A_{w^*} \gets 0$
\For{$t = 0, 1, \ldots, T-1$}
    \If{$t < T_i$}
        % \State $(z_{t-1}, _) \gets DM(z_t, \mathcal{P}, t, s)$
        \State $[z_{t}, (z^{*}_{t}, A_t)] \gets \textsc{DM}([z_t, z^*_t], [\mathcal{P}, \mathcal{P}^*], t, s)\{K^{vl}_{trg}, V^{vl}_{trg} \leftarrow K^{vl}_{src}, V^{vl}_{src}\}$
        \If{$T_i-3 \le t < T_i$}
            \State $A_{w^*} = A_{w^*} + mean(A_t[len(\mathcal{P}^*_{\text{embed}}):, \text{idx}_{w^*}], dim=1)$
        \EndIf
        \If{$t == T_i-1$}
            \State $A_{w^*} =\textsc{Normalize}(A_{w^*})$
            \State $M \gets \textsc{Preprocess}(A_{w^*})$
        \EndIf
    \ElsIf{$T_i \leq t < T_e$}        
        \State $[z_{t}, z^{*}_{t}] \gets \textsc{DM}([z_t, z^*_t], [\mathcal{P}, \mathcal{P}^*], t, s)\{K^{vl}_{trg}, V^{vl}_{trg} \leftarrow \textsc{WS}(K^{vl}_{src}, V^{vl}_{src}, K^{vl}_{trg}, V^{vl}_{trg}, M)\}$
    \Else
        \State $[z_{t}, z^{*}_{t}] \gets \textsc{DM}([z_t, z^*_t], [\mathcal{P}, \mathcal{P}^*], t, s)$
    \EndIf    
\EndFor
\State $\mathcal{V}_{\text{src}}, \mathcal{V}_{\text{trg}} \leftarrow \mathcal{D}(z_t), \mathcal{D}(z^*_t)$ 
\State \textbf{Return} $(\mathcal{V}_{\text{src}}, \mathcal{V}_{\text{trg}}))$
\end{algorithmic}
\label{alg:objadd}
\end{algorithm}

% remap:
%   set1: # best
%     k: 10.0
%     threshold: 0.1
%     apply_gaussian_blur: true
%     gaussian_kernel_size: 3
%     gaussian_sigma: 1.0
%     remap_binary_threshold: 0.8

Our general algorithm for object addition for the video $\mathcal{V}_{\text{src}}$, initially generated from a prompt $\mathcal{P}$ and seed $s$, is outlined in~\Cref{alg:objadd}. 
To generate the edited video $\mathcal{V}_{trg}$ based on a target prompt $\mathcal{P}^*$, we perform the iterative diffusion process concurrently for both the source and target prompts.

At each diffusion step $t$, we compute the attention map $A_t$ by averaging over all heads in the multi-head attention mechanism.
To extract a mask, we focus on the attention map $A_{w^*}$ corresponding to the delta words $w^*$ in the target prompt.
Prior to mask extraction, we apply the \textsc{Preprocess} function to $A_{w^*}$, which includes normalization, Gaussian blurring, and thresholding.
The resulting binary mask is then used to constrain the editing region in the denoising process.
Throughout all experiments, we used a fixed set of hyperparameters for consistency.

\subsection{Non-Rigid Editing}

\begin{algorithm}[h!]
\caption{~\modelname~Video Editing - Non-Rigid Editing}
\begin{algorithmic}[1]
\State \textbf{Input:} A source prompt $\mathcal{P}$, a target prompt $\mathcal{P}^*$, and a random seed $s$.
\State \textbf{Output:} A source video $\mathcal{V}_{\text{src}}$ and an edited video $\mathcal{V}_{\text{trg}}$.
\State $z_0 \sim \mathcal{N}(0, I)$ \Comment{a unit Gaussian random variable with random seed $s$}
\State $z^*_0 \gets z_0$
\For{$t = 0, 1, \ldots, T-1$}
    \If{$t < T_e$}        
        \State $[z_{t}, z^{*}_{t}] \gets \textsc{DM}([z_t, z^*_t], [\mathcal{P}, \mathcal{P}^*], t, s)\{K^{nl}_{trg}, V^{nl}_{trg} \leftarrow K^{nl}_{src}, V^{nl}_{src}\}$        
    \Else
        \State $[z_{t}, z^{*}_{t}] \gets \textsc{DM}([z_t, z^*_t], [\mathcal{P}, \mathcal{P}^*], t, s)$
    \EndIf    
\EndFor
\State $\mathcal{V}_{\text{src}}, \mathcal{V}_{\text{trg}} \leftarrow \mathcal{D}(z_t), \mathcal{D}(z^*_t)$ 
\State \textbf{Return} $(\mathcal{V}_{\text{src}}, \mathcal{V}_{\text{trg}})$
\end{algorithmic}
\label{alg:nonrigid}
\end{algorithm}

Our general algorithm for non-rigid editing is outlined in~\Cref{alg:nonrigid}. 
Non-rigid editing injects the keys and values from the source video into those of the target video at non-vital layers during the early diffusion steps, specifically before timestep $T_e$.

\begin{figure}[ht!]    
    \centering
    \begin{minipage}{\linewidth}                \includegraphics[width=\linewidth]{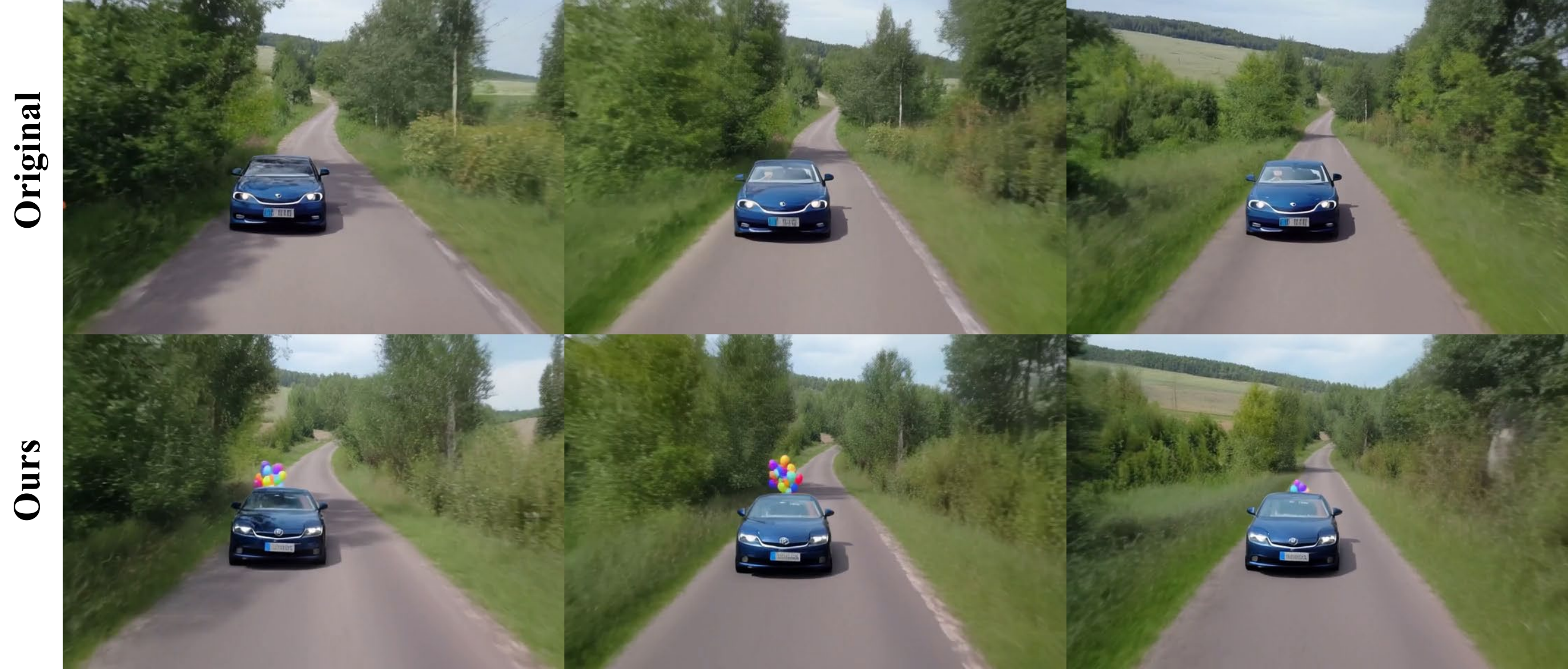}  

    \captionsetup{justification=centering} % 캡션 정렬 설정
    \caption*{\textbf{(a)} \textbf{Source prompt}: A car is driving down a country road. \\ \textbf{Target prompt}: A car driving down a country road with colorful balloons tied to it.}
    
    \end{minipage}
    
    \begin{minipage}{\linewidth}                   \includegraphics[width=\linewidth]{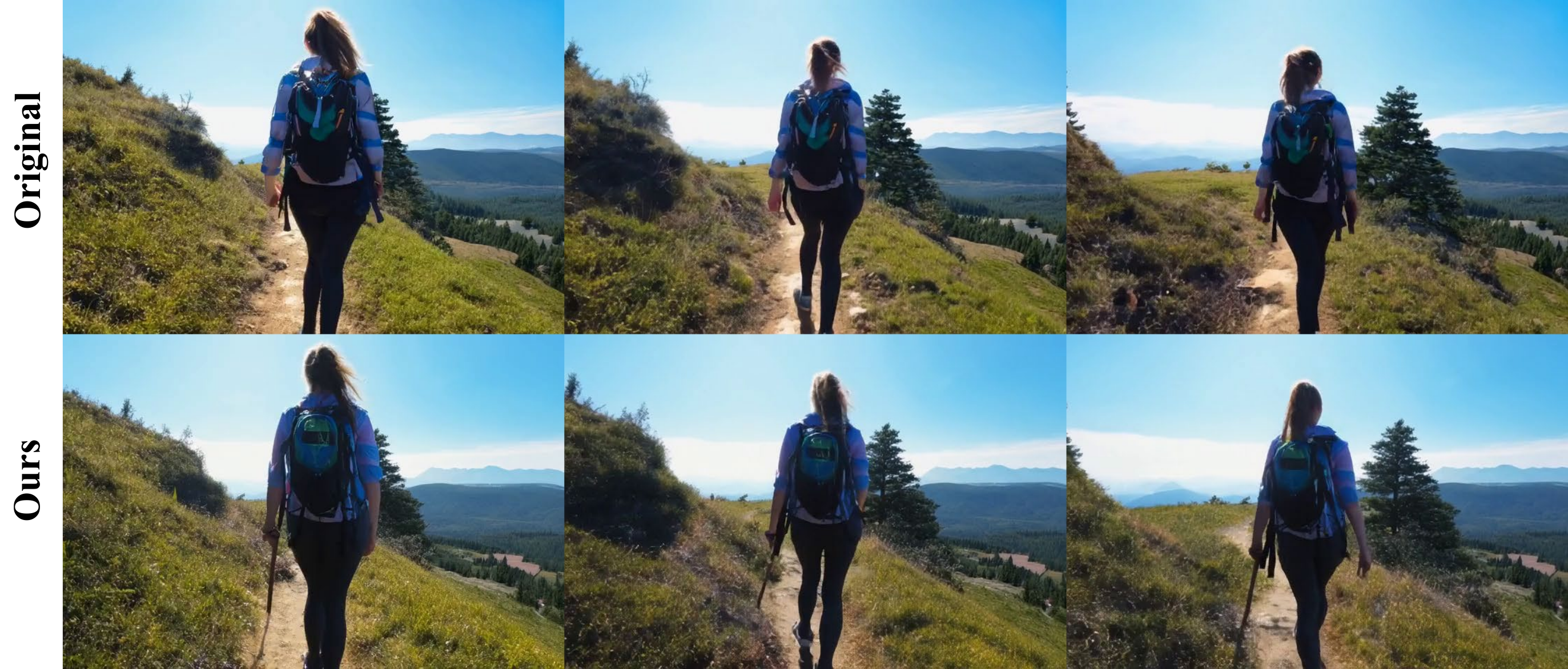}

    \captionsetup{justification=centering} % 캡션 정렬 설정
    \caption*{\textbf{(b)} \textbf{Source prompt}: A woman is hiking on a mountain trail. \\ \textbf{Target prompt}: A woman is hiking on a mountain trail carrying a walking stick.}
    \end{minipage}

    \begin{minipage}{\linewidth}                   \includegraphics[width=\linewidth]{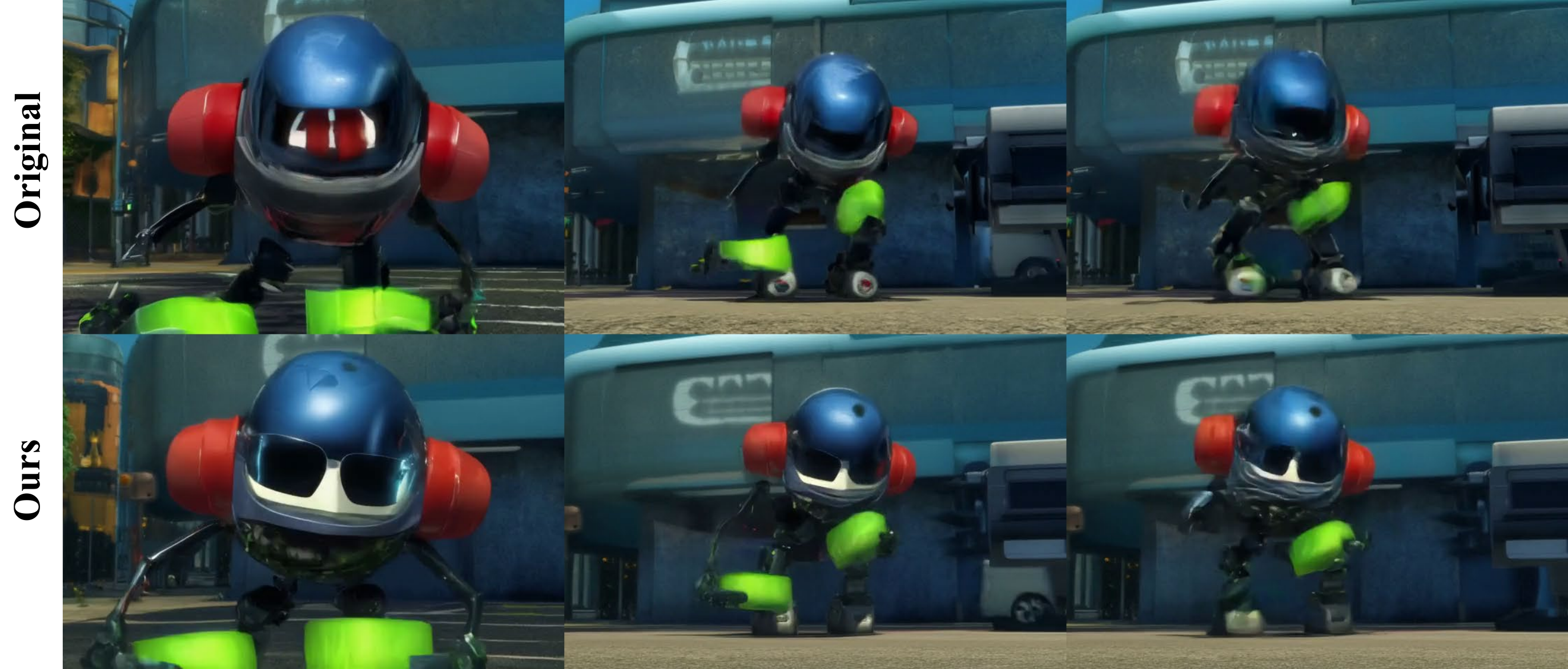}

    \captionsetup{justification=centering} % 캡션 정렬 설정
    \caption*{\textbf{(c)} \textbf{Source prompt}: A robot is standing in the middle of a city street. \\ \textbf{Target prompt}: A robot wearing sunglasses is standing in the middle of a city street.}
    \end{minipage}
    
    \caption{\textbf{Additional Object Addition Results.} Our method can insert objects into the source video without temporal inconsistency or visual artifacts.
    }  % 이미지 캡션
    \label{fig:more_objadd}
\end{figure}

\begin{figure}[ht!]    
    \centering
    \begin{minipage}{\linewidth}                \includegraphics[width=\linewidth]{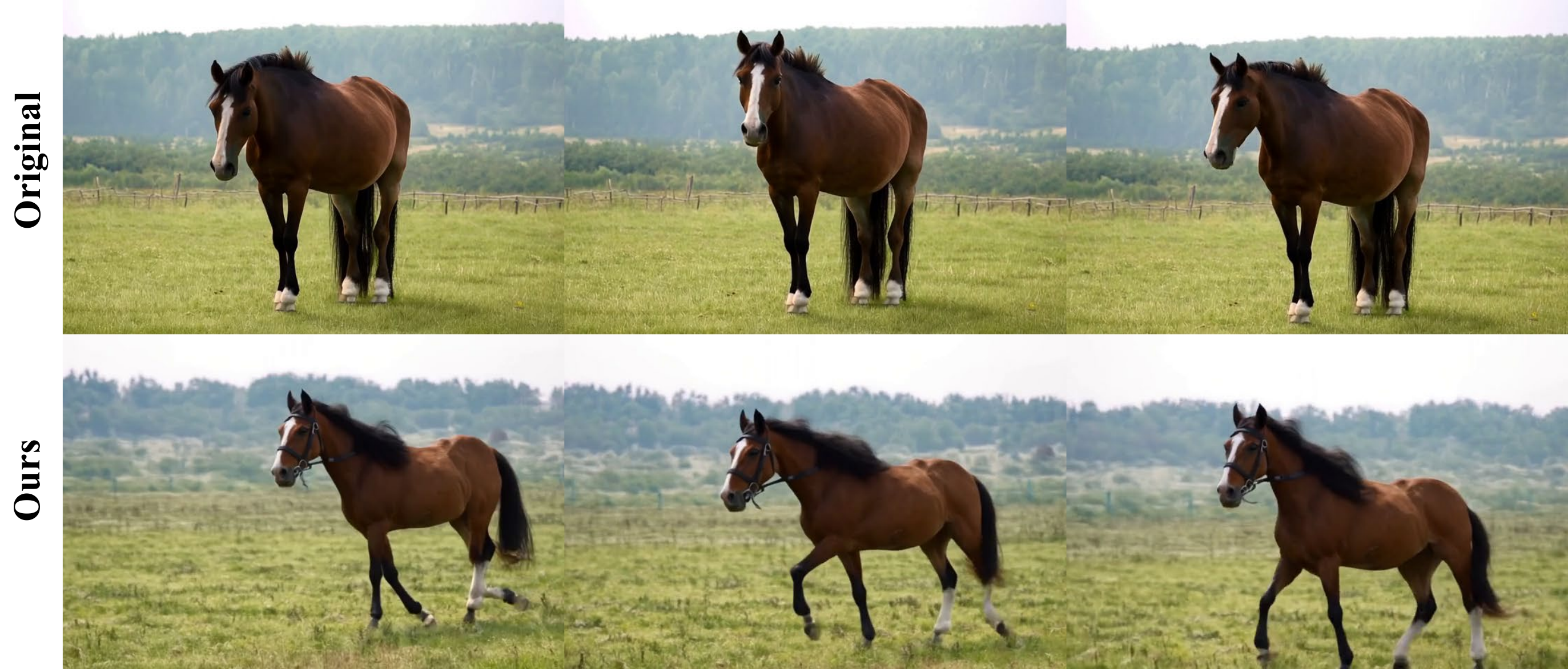}  

    \captionsetup{justification=centering} % 캡션 정렬 설정
    \caption*{\textbf{(a)} \textbf{Source prompt}: A horse standing still in a meadow. \\ \textbf{Target prompt}: A horse trotting across the field.}
    
    \end{minipage}
    
    \begin{minipage}{\linewidth}                   \includegraphics[width=\linewidth]{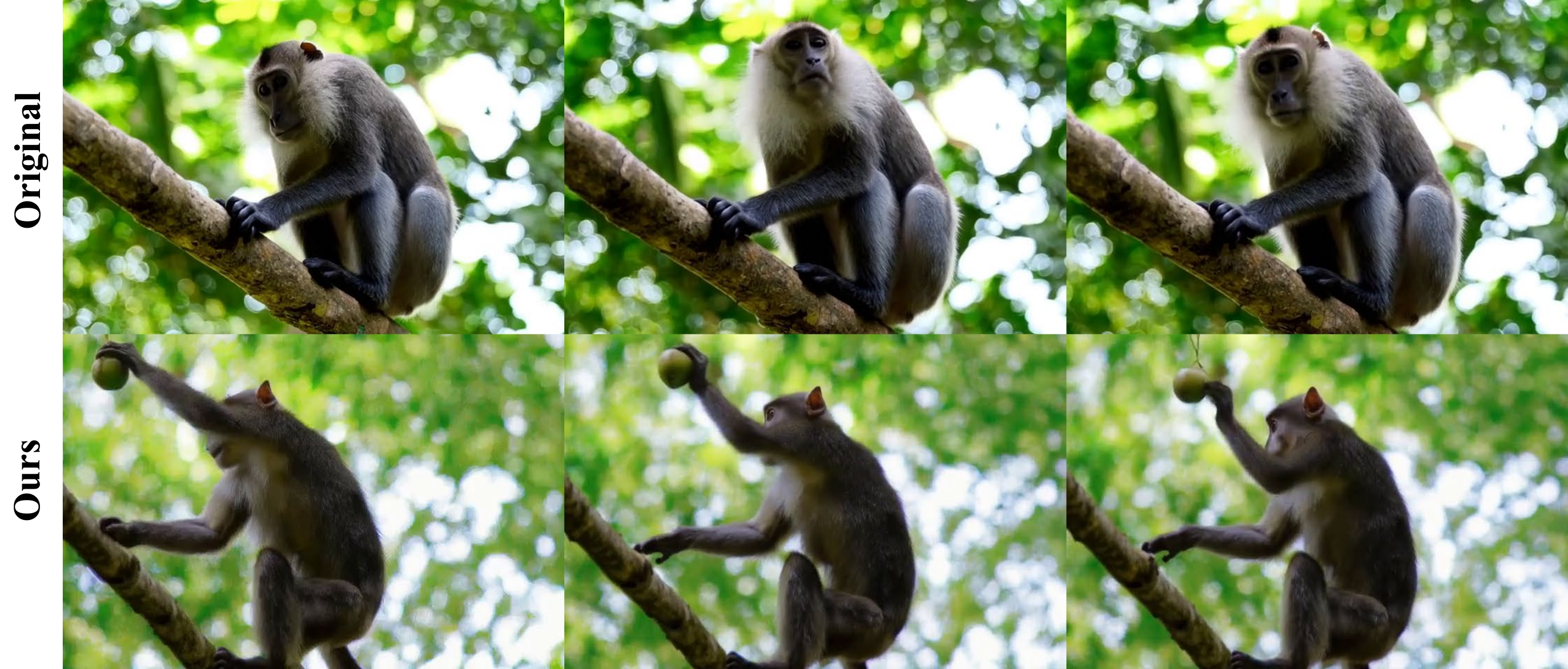}

    \captionsetup{justification=centering} % 캡션 정렬 설정
    \caption*{\textbf{(b)} \textbf{Source prompt}: A monkey sitting on a tree branch in a rainforest. \\ \textbf{Target prompt}: A monkey reaching for a fruit while on the tree branch.}
    \end{minipage}

    \begin{minipage}{\linewidth}                   \includegraphics[width=\linewidth]{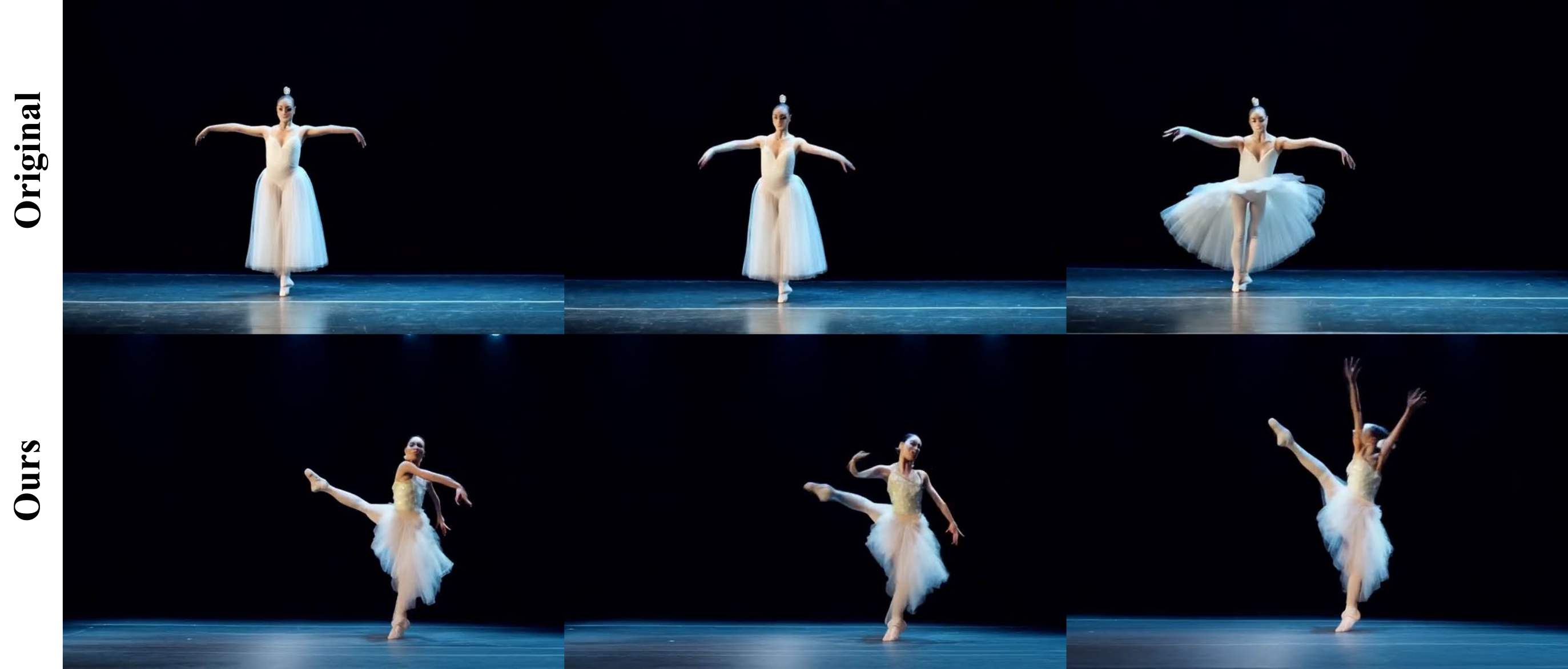}

    \captionsetup{justification=centering} % 캡션 정렬 설정
    \caption*{\textbf{(c)} \textbf{Source prompt}: A ballerina standing on a dark stage. \\ \textbf{Target prompt}: A ballerina extending one leg gracefully on the dark stage.}
    \end{minipage}
    
    \caption{\textbf{Additional Non-Rigid Editing Results.} Our method modifies the pose or motion based on the target prompt while faithfully preserving the appearance details from the source video.
    }  % 이미지 캡션
    \label{fig:more_nonrigid}
\end{figure}

\begin{figure}[ht!]    
    \centering
    \begin{minipage}{\linewidth}                \includegraphics[width=\linewidth]{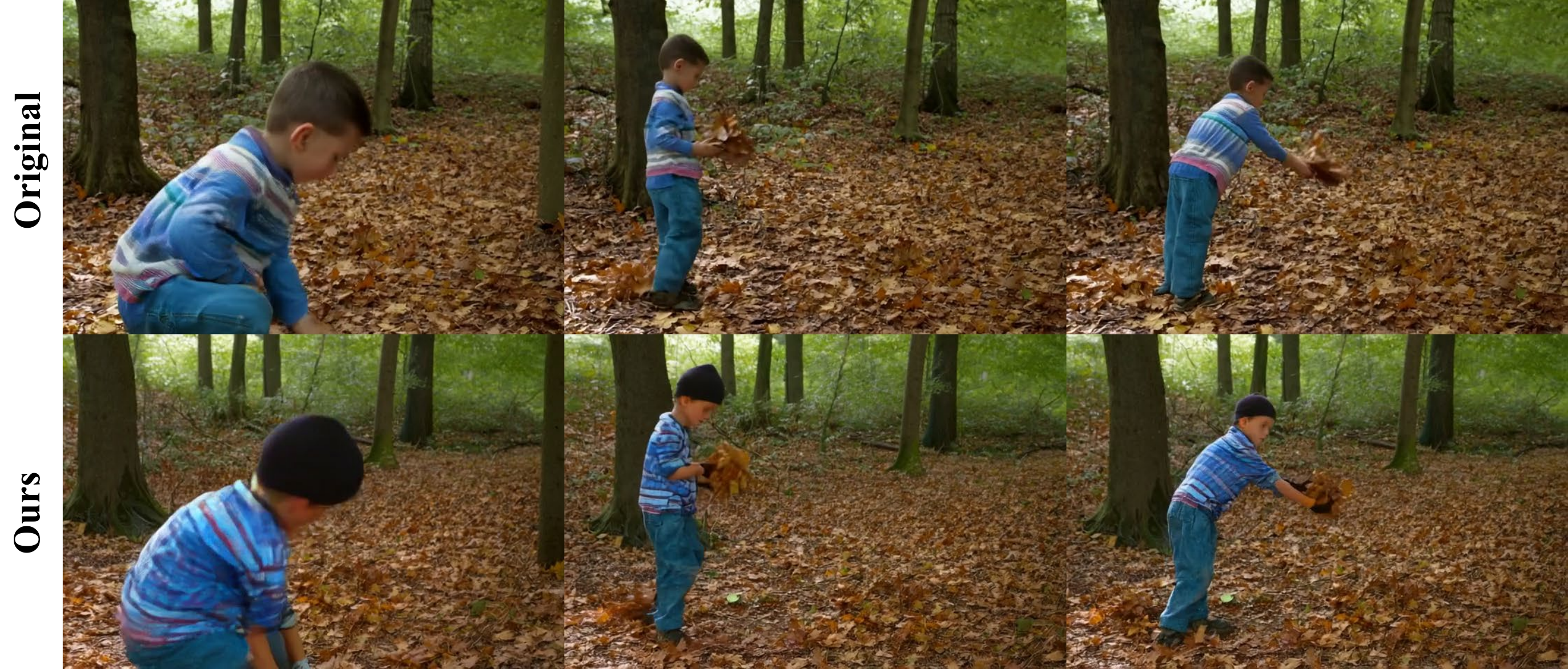}  

    \captionsetup{justification=centering} % 캡션 정렬 설정
    \caption*{\textbf{(a)} \textbf{Source prompt}: A boy is collecting leaves in the forest. \\ \textbf{Target prompt}: A boy wearing a beanie and gloves is collecting leaves in the forest.}
    
    \end{minipage}
    
    \begin{minipage}{\linewidth}                   \includegraphics[width=\linewidth]{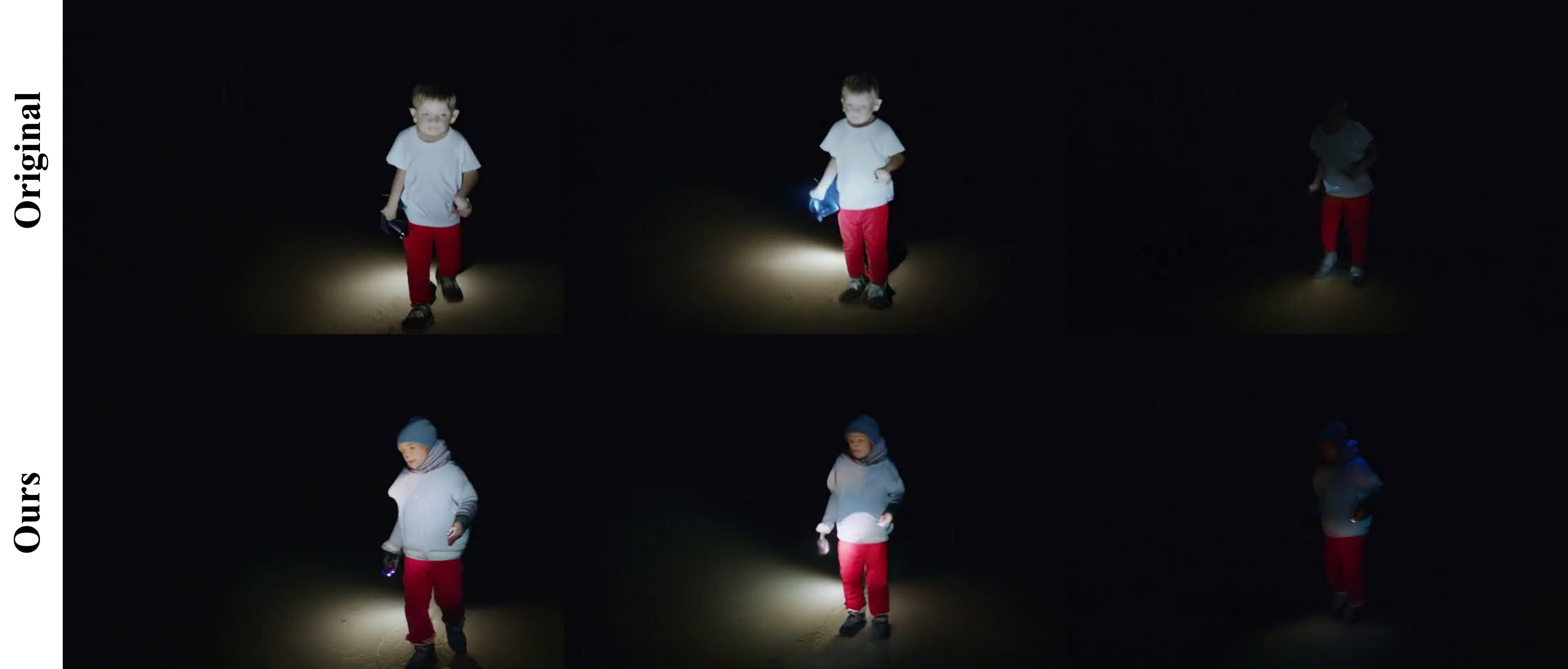}

    \captionsetup{justification=centering} % 캡션 정렬 설정
    \caption*{\textbf{(b)} \textbf{Source prompt}: A boy is walking with a flashlight. \\ \textbf{Target prompt}: A boy wearing a scarf and beanie is walking with a flashlight.}
    \end{minipage}

    \begin{minipage}{\linewidth}                   \includegraphics[width=\linewidth]{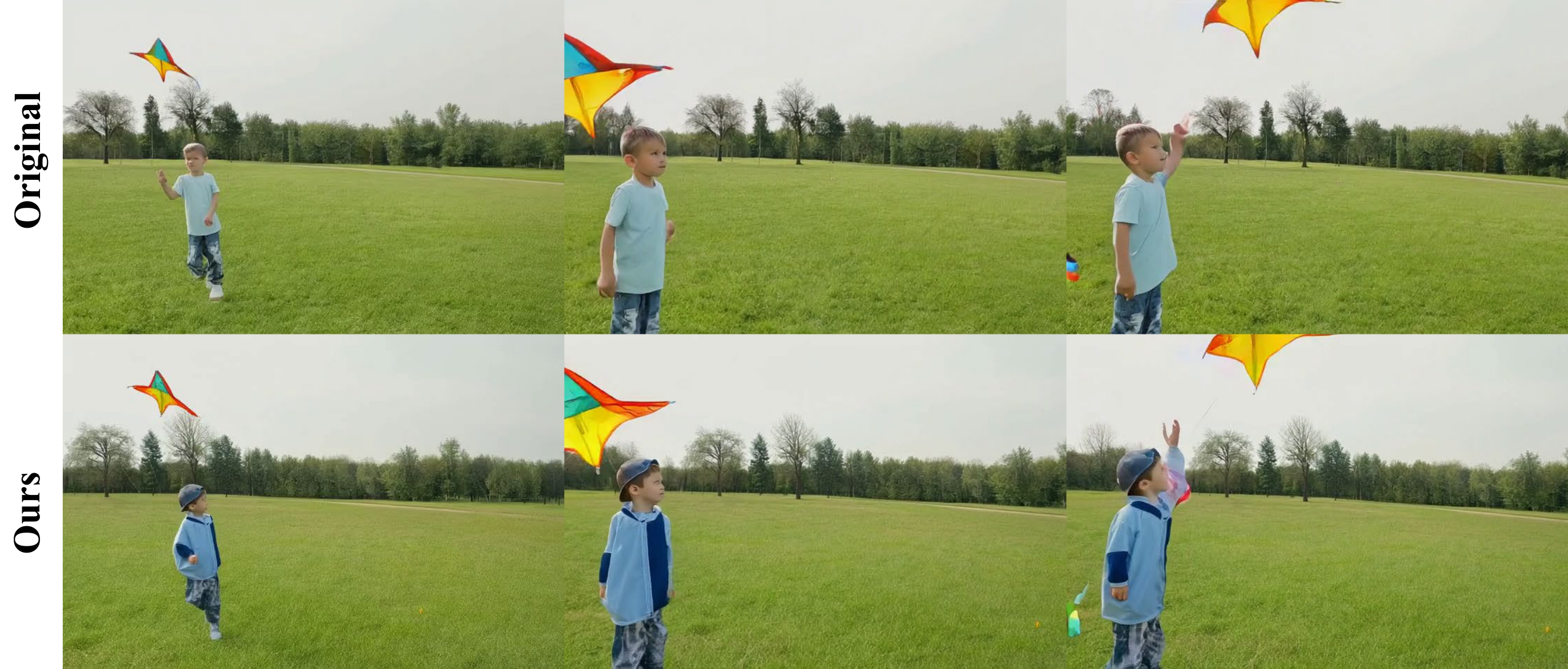}

    \captionsetup{justification=centering} % 캡션 정렬 설정
    \caption*{\textbf{(c)} \textbf{Source prompt}: A boy is flying a kite in the park. \\ \textbf{Target prompt}: A boy wearing a windbreaker and a cap is flying a kite in the park.}
    \end{minipage}
    
    \caption{\textbf{Multi-Object Addition Results.} Our method can insert multiple objects by simply appending additional words to the target prompt.
    }  % 이미지 캡션
    \label{fig:more_multi}
\end{figure}

\section{Additional Qualitative Results}
In this section, we present additional qualitative results. 
\Cref{fig:more_objadd} shows further examples of object addition, while \Cref{fig:more_nonrigid} illustrates additional results for non-rigid editing. 
As demonstrated in \Cref{fig:more_multi}, our method also supports multi-object addition by simply extending the target prompt with additional words.

Video results are available at the anonymous link below: \\
\url{https://anonymous.4open.science/w/TV_LiVE-71C4/}

\begin{figure}[ht!]    
    \centering
    \begin{minipage}{\linewidth}                \includegraphics[width=\linewidth]{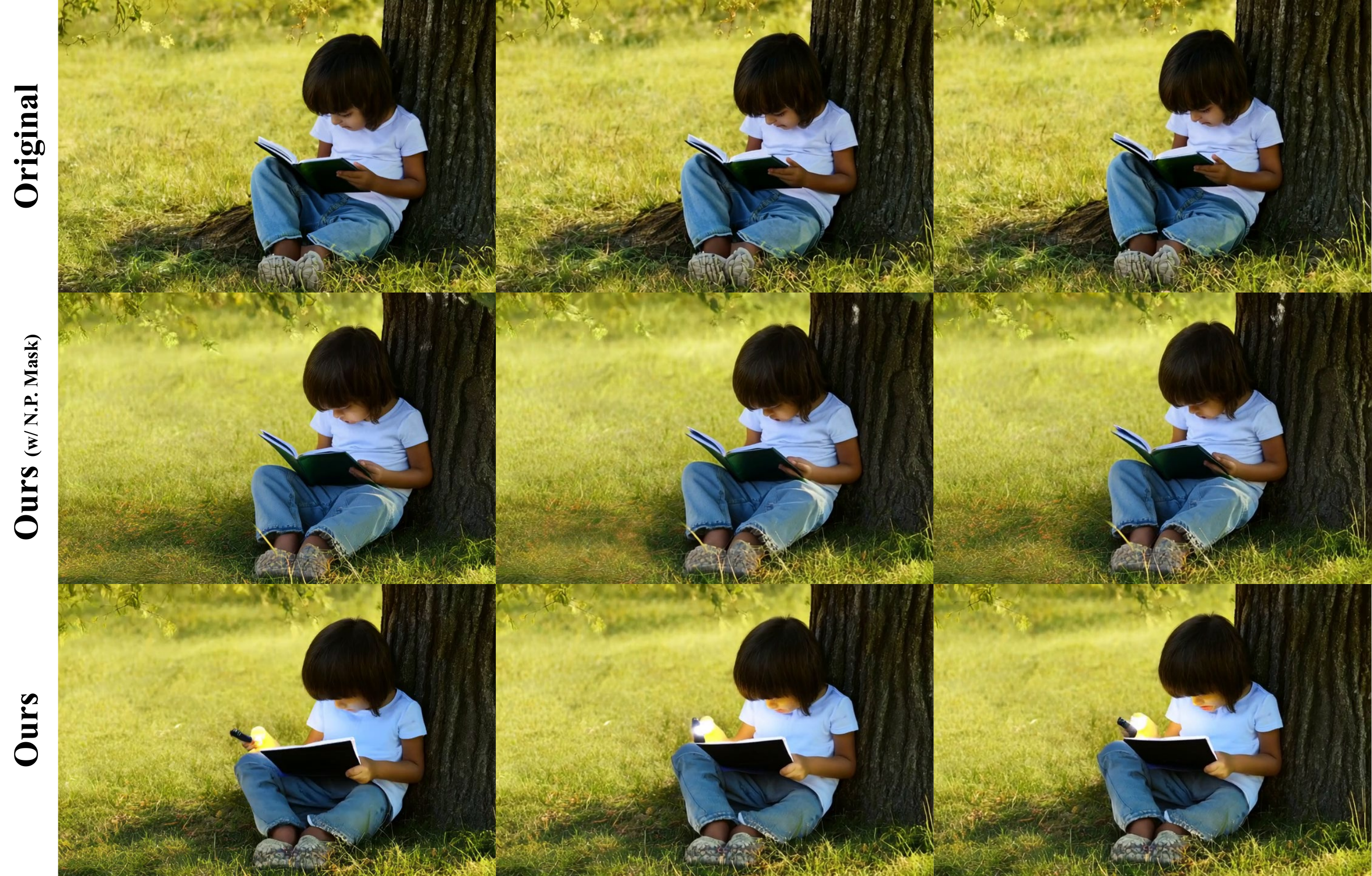}  

    \captionsetup{justification=centering} % 캡션 정렬 설정
    \caption*{\textbf{(a)} \textbf{Source prompt}: A child is reading under a tree. \\ \textbf{Target prompt}: A child with a flashlight is reading under a tree.}
    
    \end{minipage}
    \begin{minipage}{\linewidth}                   \includegraphics[width=\linewidth]{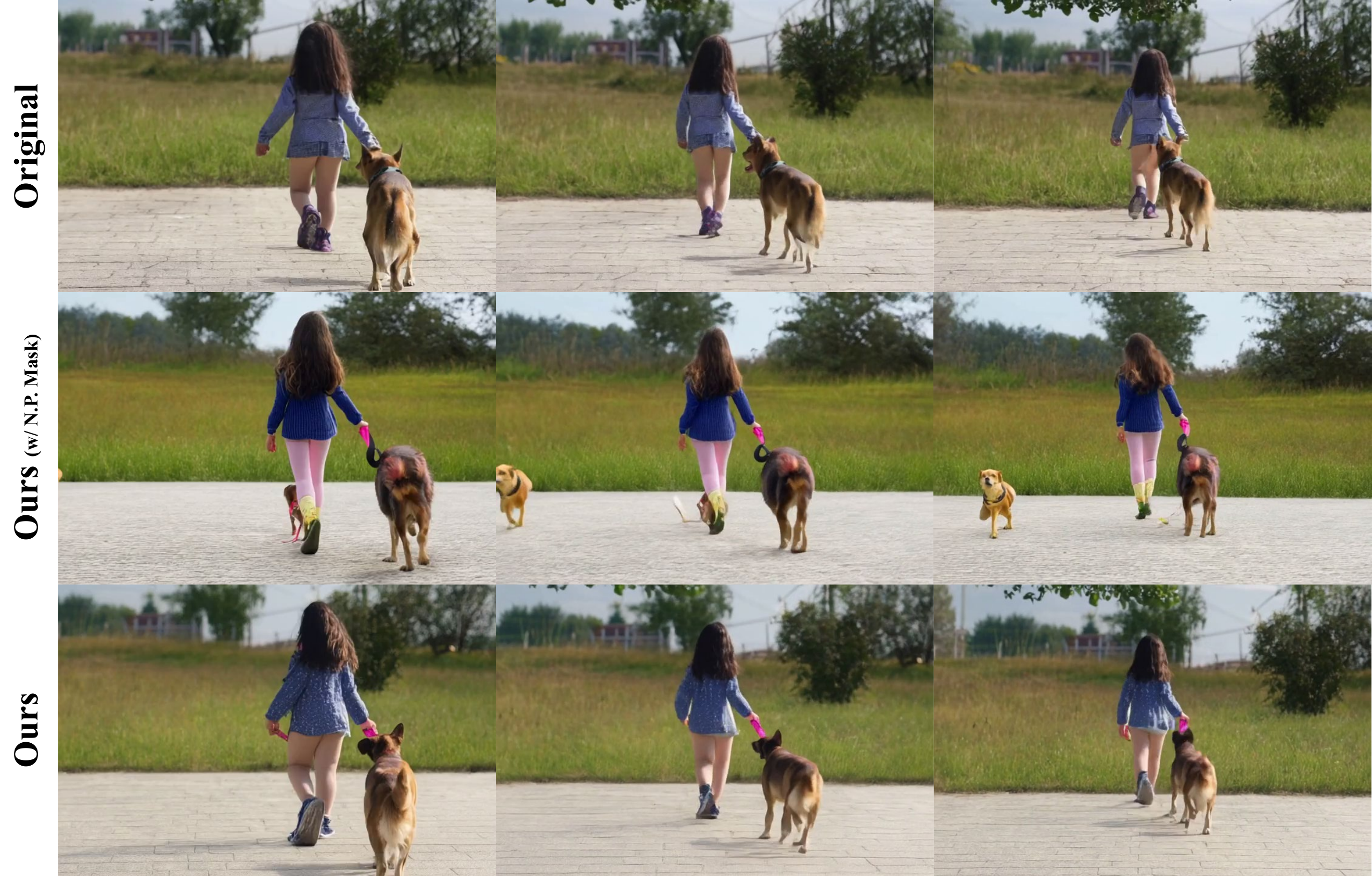}

    \captionsetup{justification=centering} % 캡션 정렬 설정
    \caption*{\textbf{(b)} \textbf{Source prompt}: A girl is walking her dog. \\ \textbf{Target prompt}: A girl with a pink leash is walking her dog.}
    \end{minipage}
    
    \caption{\textbf{Prominent Layer Ablation.} Using masks extracted from non-prominent layers for object addition often results in (a) failure to add the target object or (b) excessive and unintended modifications across the entire video, highlighting the importance of selecting the appropriate layer for mask extraction.
    }  % 이미지 캡션
    \label{fig:ablation_rigid}
\end{figure}

\begin{figure}[ht!]    
    \centering
    \begin{minipage}{\linewidth}                \includegraphics[width=\linewidth]{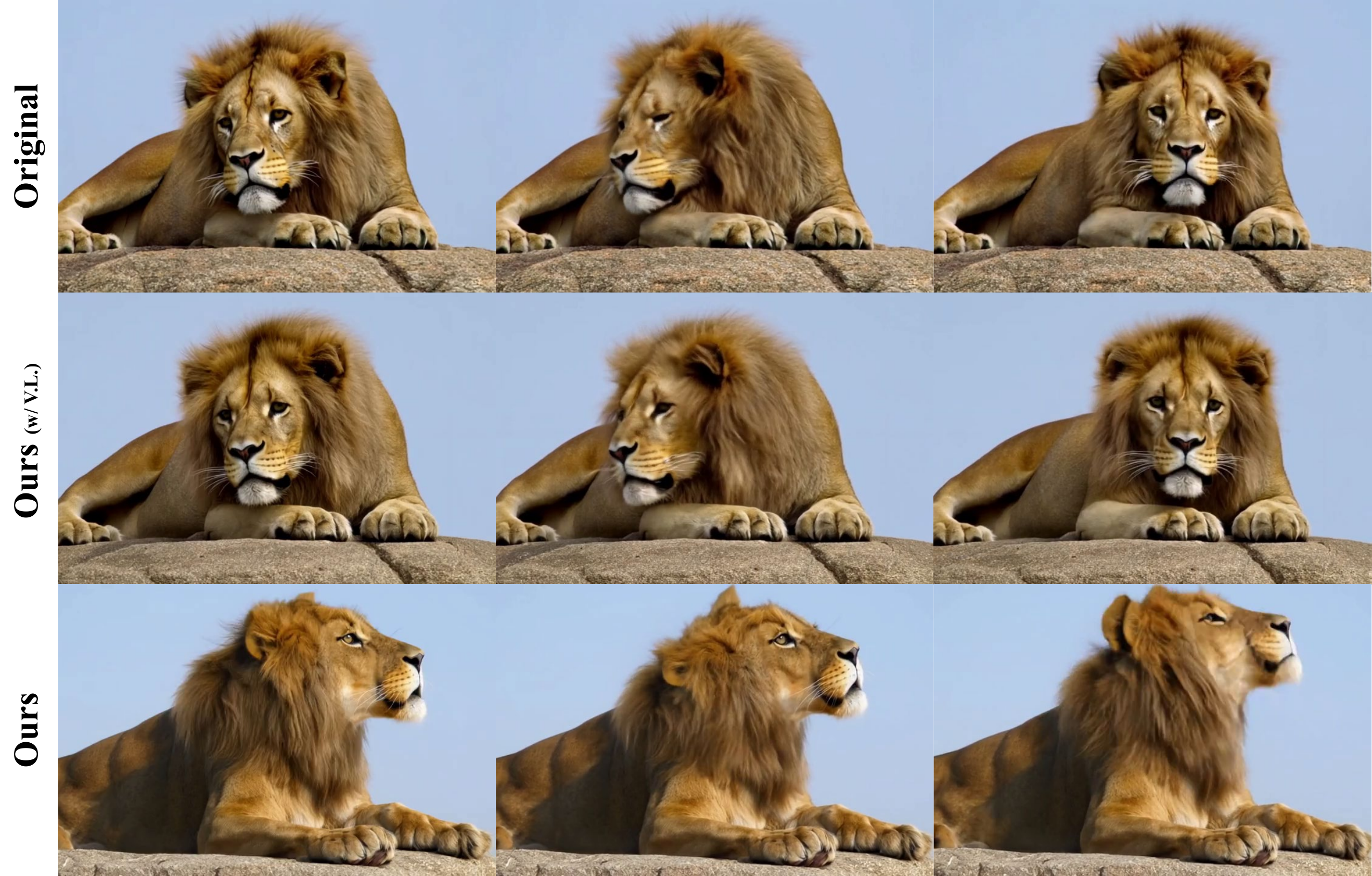}   

    \captionsetup{justification=centering} % 캡션 정렬 설정
    \caption*{\textbf{(a)} \textbf{Source prompt}: a lion lying on a rock.\\ \textbf{Target prompt}: a lion lifting its head while lying on the rock.}
    
    \end{minipage}
    \begin{minipage}{\linewidth}                   \includegraphics[width=\linewidth]{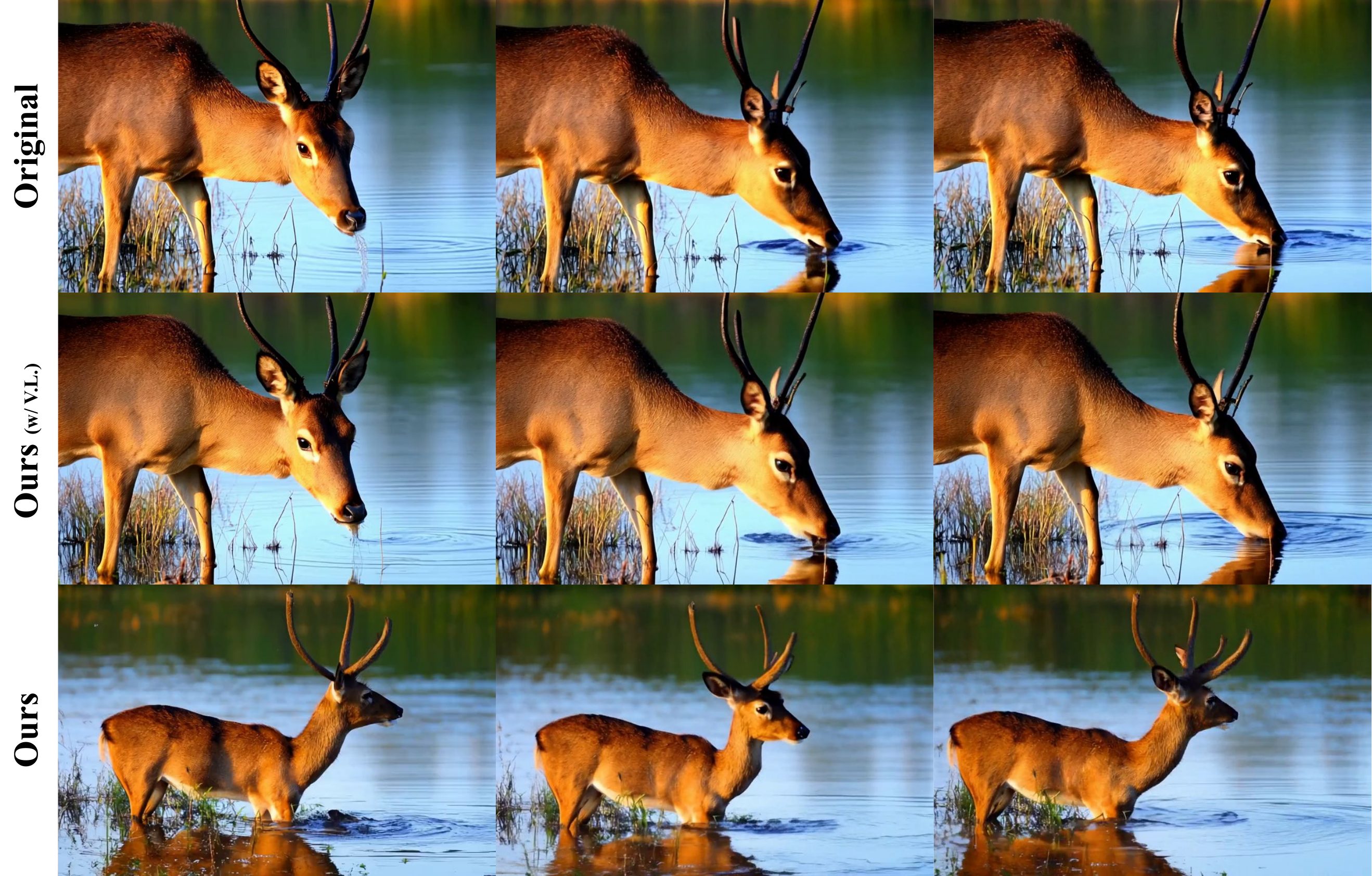}   

    \captionsetup{justification=centering} % 캡션 정렬 설정
    \caption*{\textbf{(b)} \textbf{Source prompt}: a deer drinking water from a lake at dawn. \\ \textbf{Target prompt}: a deer raising its head alertly from the lake.}
    \end{minipage}
    
    \caption{\textbf{Vital Layer Ablation.} Using non-vital layers for non-rigid editing results in no changes to the output, showing the importance of using non-vital layers for non-rigid editing.}  % 이미지 캡션
    \label{fig:ablation_nonrigid}
\end{figure}

\section{Ablation Study Visualization}
In the main paper (see~\Cref{sec:quan}), we conducted two ablation studies.
The corresponding quantitative results are reported in~\Cref{tab:quant}.
Here, we provide a qualitative comparison. 

\subsection{Object Addition: Prominent Layer Ablation}

The first study investigates the impact of not using prominent layers. 
As described in~\Cref{sec:obj_add}, our method performs object addition by utilizing masks extracted from prominent layers, which are identified based on the proposed Layer Prominence metric.
For this ablation, we instead used masks extracted from the layer with the lowest prominence value, which corresponds to layer 33. 
In \Cref{fig:ablation_rigid}, when masks from non-prominent layers are used, the model often fails to add the target object or introduces excessive and unintended changes across the entire video. 
In contrast, our proposed method, \modelname, successfully modifies only the relevant regions of the video while preserving the rest, demonstrating its effectiveness in performing controlled and localized edits.

\subsection{Non-Rigid Editing: Vital Layer Ablation}

The second study investigates the impact of using vital layers for non-rigid video editing. 
As described in~\Cref{sec:non_rigid}, our method performs non-rigid editing by utilizing non-vital layers, which are identified using the RoPE Vitality metric.
For this ablation, we instead used layers with high RoPE vitality values (i.e., vital layers). 
As shown in~\Cref{fig:ablation_nonrigid}, when vital layers are used, the model fails to apply any meaningful edit and merely replicates the source video. 
This behavior is consistent with our earlier findings in~\Cref{sec:vitality}, where we observed that injecting information from vital layers alone is insufficient for object addition, often resulting in no changes to the output.
In contrast, our proposed method, \modelname, successfully modifies only the target motion or pose of the source video while preserving the object's original appearance.
This highlights the importance of selecting appropriate layers based on RoPE vitality for successful non-rigid editing.

\section{Additional Ablation Study}
\begin{table*}[h!]
\captionsetup{type=table}
\caption{Ablation on target prompt updating. We multiply \textbf{CLIP$_{dir}$}, \textbf{CLIP$_{img}$}, Temporal Flickering (\textbf{T.F.}), Motion Smoothness (\textbf{M.S.}), Subject Consistency (\textbf{S.C.}), and Background Consistency (\textbf{B.C.}) to compute the overall evaluation score. Best results are shown in \textbf{bold}.}

\centering
\resizebox{\textwidth}{!}{%
\begin{tabular}{ll|cc|c|cccc|c}
\toprule
\textbf{Task} & \textbf{Method} & \textbf{CLIP$_{dir}$~$\uparrow$} & \textbf{CLIP$_{img}$~$\uparrow$} & \textbf{CLIP$_{all}$~$\uparrow$} & \textbf{T.F.~$\uparrow$} & \textbf{M.S.~$\uparrow$} & \textbf{S.C.~$\uparrow$} & \textbf{B.C.~$\uparrow$} & \textbf{Overall~$\uparrow$} \\
\midrule

\multirow{2}{*}{\textbf{Object Addition}} 
& \modelname{} (\text{\scriptsize w/ U.P.}) & 0.0931 & \textbf{0.9341} & 0.2643 & 0.9659 & 0.9804 & \textbf{0.9398} & \textbf{0.9474} & 0.2228\\
& \textbf{\modelname} & \textbf{0.1258} & 0.9294 & \textbf{0.2715} & \textbf{0.9660} & \textbf{0.9810} & 0.9340 & 0.9450  & \textbf{0.2286} \\
\bottomrule
\end{tabular}%
}

\label{tab:prompt_update}
\end{table*}
In CogVideoX~\cite{yang2024cogvideox}, the latent representations corresponding to the prompt and visual content are concatenated along the sequence dimension and jointly processed via self-attention. 
As a result, injecting keys and values from the source video may inadvertently update the prompt representation of the target, since both types of information share the attention space.

However, in the case of object addition, the target prompt contains information about an object that does not exist in the source video. 
Therefore, allowing the source video to influence the target prompt may lead to semantic misalignment. 
To prevent this, we mask out the attention regions corresponding to the text-to-video during inference, thereby preserving the integrity of the target prompt representation.

~\Cref{tab:prompt_update} compares performance between two settings: one where the target prompt is updated using source video features (w/ U.P.) and one where it is not (ours). 
As shown in the table, preserving the target prompt without update yields better results.

\begin{figure}[th!]    
    \centering
    \begin{minipage}{\linewidth}                \includegraphics[width=\linewidth]{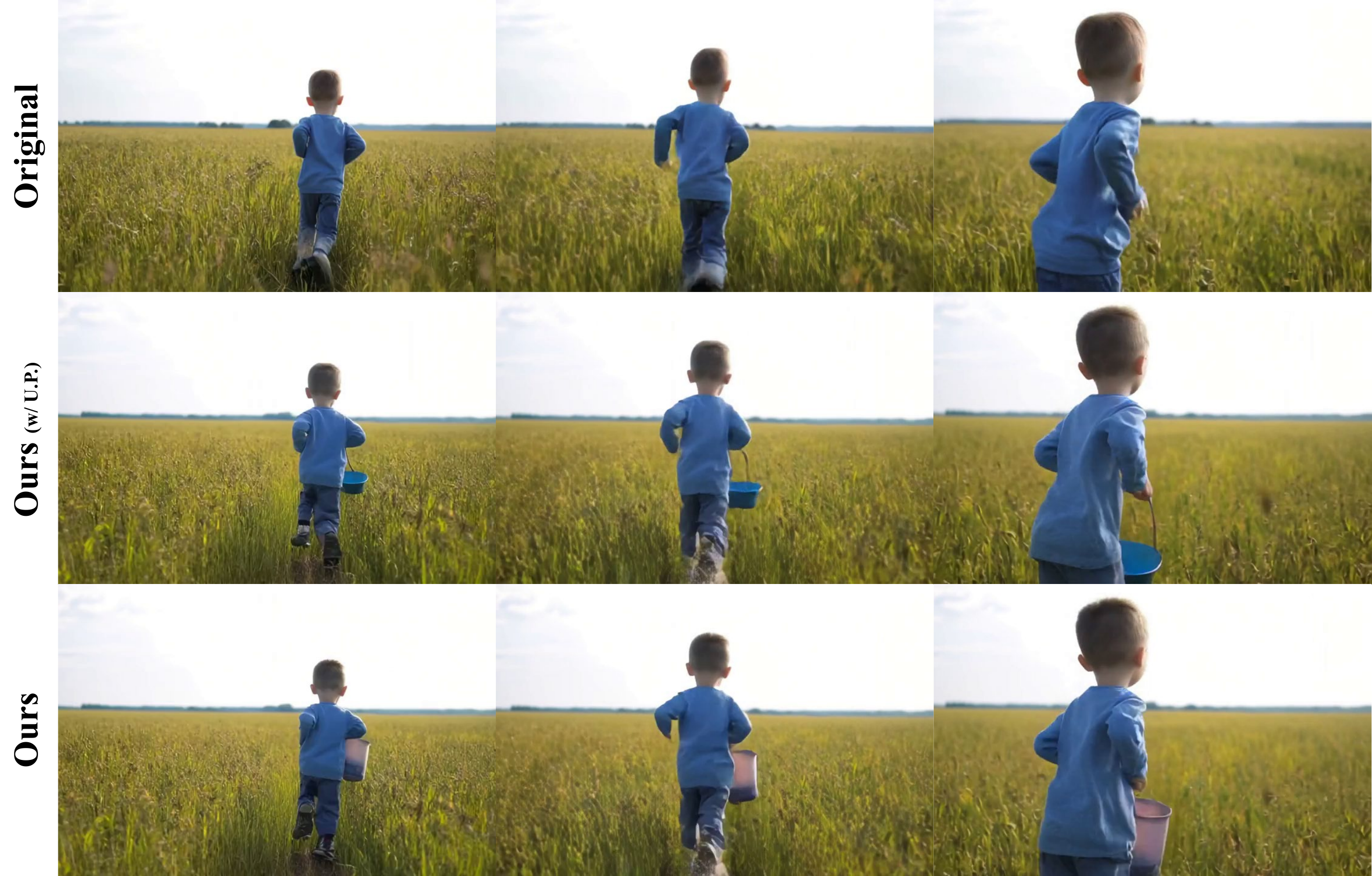}   

    \captionsetup{justification=centering} % 캡션 정렬 설정
    \caption*{\textbf{(a)} \textbf{Source prompt}: a boy running across a field.\\ \textbf{Target prompt}: a boy running across a field while holding a bucket.}
    
    \end{minipage}
    \begin{minipage}{\linewidth}                   \includegraphics[width=\linewidth]{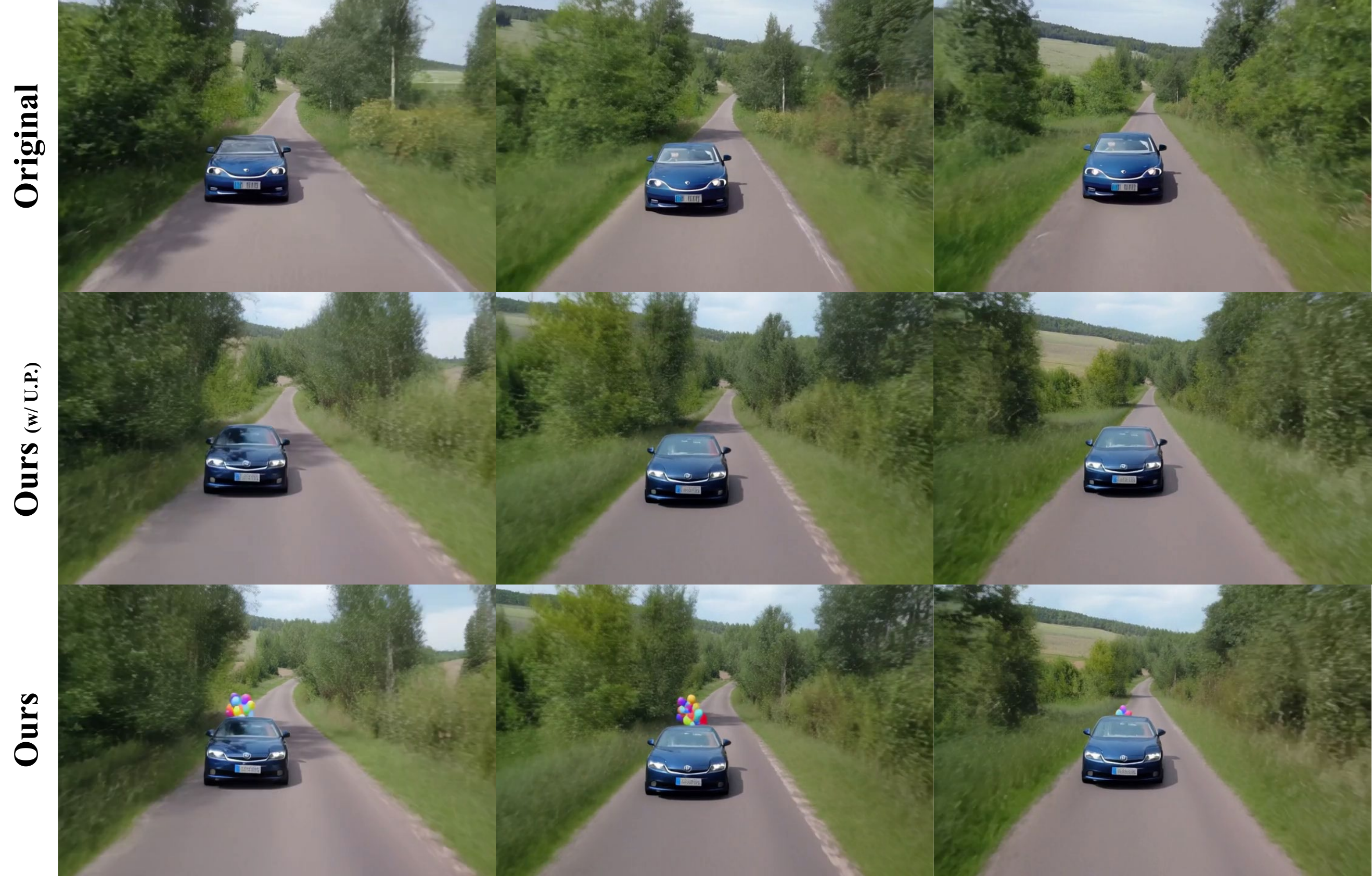}   

    \captionsetup{justification=centering} % 캡션 정렬 설정
    \caption*{\textbf{(b)} \textbf{Source prompt}: A car is driving down a country road. \\ \textbf{Target prompt}: A car driving down a country road with colorful balloons tied to it.}
    \end{minipage}
    
    \caption{
        \textbf{Ablation on Target Prompt Update.} Updating the target prompt using the source video during object addition (a) generally preserves the intended edit, but (b) can sometimes lead to no visible change, suggesting that the target prompt is being overridden by information from the source video.
    }  % 이미지 캡션
    \label{fig:ablation_update_prompt}
\end{figure}

~\Cref{fig:ablation_update_prompt} illustrates the effect of updating the target prompt using source video features during object addition. 
While this approach often maintains the intended edits, it can occasionally lead to no noticeable changes in the output, indicating that the target prompt may be suppressed by the information from the source video.

\section{User Study Design and Prompts}
To evaluate our method on \textit{object addition} and \textit{non-rigid editing} tasks, we conducted a user study comparing six video editing approaches: BivDiff~\cite{shi2024bivdiff}, RAVE~\cite{kara2024rave}, VidToMe~\cite{li2024vidtome}, CogInv~\cite{yang2024cogvideox}, CogV2V~\cite{yang2024cogvideox}, and ~\modelname (Ours).

For each task, we prepared 15 examples. Each example includes a source–target prompt pair, the corresponding source video, and six edited videos generated from the target prompt, one for each method.

Participants were asked to answer three questions for every example, selecting the method they preferred most. 
The results reported in the main paper (\Cref{tab:user}) represent the percentage of participants who selected each method as their top preference.

\subsection{Object Addition Evaluation}

For the object addition task, participants were asked the following three questions:
\begin{enumerate}
    \item Please select the video you prefer the most based on how well the target object appears and integrates with the existing elements in the scene.
    \item Please select the video you prefer the most based on how well the source video is preserved in regions other than the target object, while successfully incorporating the new content.
    \item Please select the video you would prefer the most if you were the user of this video editing tool.
\end{enumerate}

The following 15 source–target prompt pairs were used for the object addition evaluation:

\begin{itemize}
    \item Source: a boy running across a field\\
          Target: a boy running across a field \textit{while holding a bucket}
    \item Source: A girl is playing in a snowy park\\
          Target: A girl \textit{with a yellow scarf} is playing in a snowy park
    \item Source: a person standing still in a snowy landscape\\
          Target: a person standing still in a snowy landscape \textit{holding a snowboard}
    \item Source: A side view of a child painting at an easel\\
          Target: A side view of a child painting at an easel \textit{wearing a colorful apron}
    \item Source: A turtle is walking on the sand\\
          Target: A turtle \textit{with a leaf on its back} is walking on the sand
    \item Source: A child is reading under a tree\\
          Target: A child \textit{with a flashlight} is reading under a tree
    \item Source: A boy is walking through a grassy field\\
          Target: A boy \textit{holding a red balloon} is walking through a grassy field
    \item Source: A reindeer is strolling in the forest\\
          Target: A reindeer \textit{with flower crown} is strolling in the forest
    \item Source: A side view of a dog sitting on the beach\\
          Target: A side view of a dog \textit{wearing sunglasses} sitting on the beach
    \item Source: A penguin is waddling on icy ground\\
          Target: A penguin \textit{wearing a green scarf} is waddling on icy ground
    \item Source: A car is driving down a country road\\
          Target: A car driving down a country road \textit{with colorful balloons tied to it}
    \item Source: A dog is chasing a ball in a park\\
          Target: A dog \textit{wearing a green collar} is chasing a ball in a park
    \item Source: A woman is hiking on a mountain trail\\
          Target: A woman is hiking on a mountain trail \textit{carrying a walking stick}
    \item Source: A robot is standing in the middle of a city street\\
          Target: A robot \textit{wearing sunglasses} is standing in the middle of a city street
    \item Source: A dog is lying on a blanket\\
          Target: A dog \textit{wearing a Christmas sweater} is lying on a blanket
\end{itemize}

\subsection{Non-Rigid Editing Evaluation}

For the non-rigid editing task, participants were asked to answer the following three questions:
\begin{enumerate}
    \item Please select the video you prefer the most based on how accurately the target motion or pose is reflected.
    \item Please select the video you prefer the most based on how well the appearance of the source video is preserved while reflecting the target prompt.
    \item Please select the video you would prefer the most if you were the user of this video editing tool.
\end{enumerate}

The following 15 source–target prompt pairs were used for the non-rigid editing evaluation:

\begin{itemize}
    \item Source: a lamb resting in a meadow\\
          Target: a lamb resting in a meadow \textit{while rolling over onto its side}
    \item Source: a horse standing still in a meadow\\
          Target: a horse \textit{trotting across the field}
    \item Source: a bird perched on a wet branch\\
          Target: a bird \textit{flapping its wings} while perched on a wet branch
    \item Source: a skier paused on a mountain slope during a light snowfall\\
          Target: a skier \textit{adjusting their goggles} on the snowy mountain slope
    \item Source: a firefighter standing in front of a burning building\\
          Target: a firefighter \textit{pointing directions to others} at the burning building
    \item Source: a wolf standing in a snowy forest\\
          Target: a wolf \textit{howling with its head tilted upward} in the snowy forest
    \item Source: a ballerina standing on a dark stage\\
          Target: a ballerina \textit{extending one leg gracefully} on the dark stage
    \item Source: a monkey sitting on a tree branch in a rainforest\\
          Target: a monkey \textit{reaching for a fruit} while on the tree branch
    \item Source: a man standing under a tree with falling leaves\\
          Target: a man \textit{kneeling to pick up a leaf from the ground}
    \item Source: a squirrel perched on a tree branch\\
          Target: a squirrel perched on a tree branch \textit{while chewing on an acorn}
    \item Source: a deer drinking water from a lake at dawn\\
          Target: a deer \textit{raising its head alertly} from the lake
    \item Source: a cat sitting in a garden\\
          Target: a cat \textit{raising one paw} in the garden
    \item Source: a turtle resting on a rock\\
          Target: a turtle \textit{extending its neck} while resting on the rock
    \item Source: a dog running in the rain\\
          Target: a dog \textit{shaking off} the rain
    \item Source: a boy standing still in a swimming pool\\
          Target: a boy \textit{crouching slightly and touching the water} in the swimming pool
\end{itemize}

% % 1-1.  목표로 하는 물체가 등장하여 기존의 물체들과 잘 어우러졌는지를 바탕으로 가장 선호하는 동영상을 고르시오.
% Please select the video you prefer the most based on how well the target object appears and integrates with the existing elements in the scene.
% % 1-2.  Target Prompt가 반영된 영상 중, 추가된 물체 이외의 영역에서 source video를 잘 보존하였는지를 바탕으로 가장 선호하는 동영상을 고르시오. 
% Please select the video you prefer the most based on how well the source video is preserved in regions other than the target object, while adding new content.
% % 1-3. 만약 당신이 이 동영상 편집 툴의 사용자라면 어떤 동영상을 사용할 지 고르시오 
% Please select the video you would prefer the most if you were the user of this video editing tool.

% % 1-1.   목표로 하는 움직임이나 자세가 잘 반영됐는지를 바탕으로 가장 선호하는 동영상을 고르시오.
% Please select the video you prefer the most based on how accurately the target motion or pose is reflected.
% % 1-2.  Target Prompt가 반영된 영상 중,  source video에 포함된 외형 정보(객체, 배경이 되는 장소)를 얼마나 잘 반영했는지를 바탕으로 가장 선호하는 동영상을 고르시오.
% Please select the video you prefer the most based on how well the appearance of the source video is preserved while reflecting the target prompt.
% % 1-3. 만약 당신이 이 동영상 편집 툴의 사용자라면 어떤 동영상을 사용할 지 고르시오 
% Please select the video you would prefer the most if you were the user of this video editing tool.

\section{Computational Resources}
To analyze hardware resource utilization, we employed a video generation setup with a resolution of 720×480 pixels and 49 frames. 
The generation was conditioned on a source prompt “a boy running across a field” and a target prompt “a boy running across a field while holding a bucket.” 
Measurements were conducted five times to ensure consistency. 
Compared to the backbone model, our method exhibited increased processing time and memory usage, requiring an average of approximately 1206 seconds and 33.86 GB of peak memory, whereas the backbone model required around 1045 seconds and 28.91 GB of memory.

\section{Limitations}
\label{sec:limit}
\begin{figure}[t]
    \centering    
    \includegraphics[width=\linewidth]{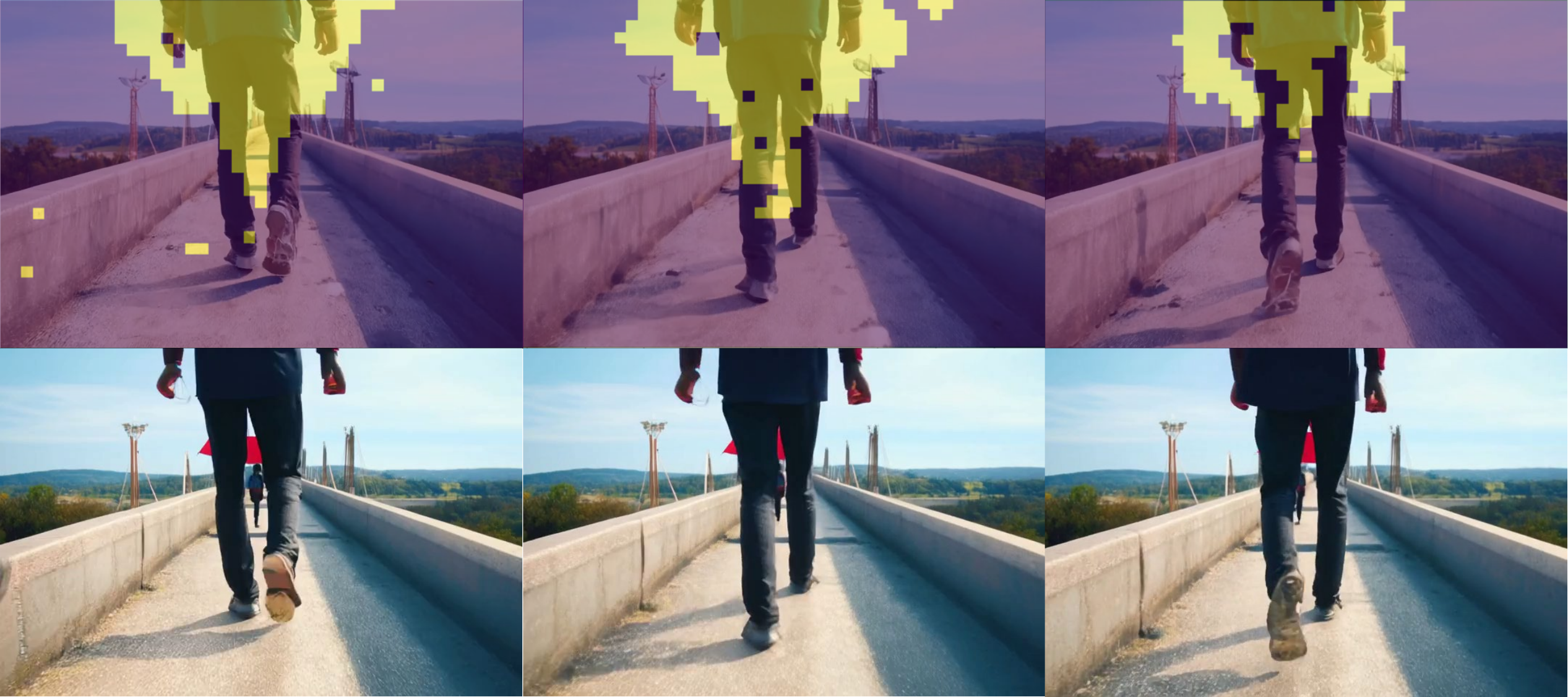}
    \caption{
    Visualization of our failure case in object addition.    
    The top row shows the sampled source video frames with the automatically extracted mask overlaid.
    The bottom row shows the resulting frames generated with the target prompt.
    In this example, the mask incorrectly highlights the background area 
    % instead of the intended region (e.g., the man's hand)
    , leading to the unintended insertion of a new person holding a red umbrella. \textbf{Source Prompt}: A man is walking on a bridge. \textbf{Target Prompt}: A man holding a red umbrella is walking on a bridge.
    }

    \label{fig:limitation_mask}
\end{figure}

\noindent \textbf{Unexpected Mask Extraction in Object Addition}\\
One limitation of our method lies in the automatic extraction of the editing mask for object addition. The mask is derived from attention maps corresponding to the delta tokens between the source and target prompts. However, this automated process may not always align with the user's intent.

For instance, in ~\Cref{fig:limitation_mask}, the prompt modification ``holding a red umbrella'' results in a mask that incorrectly highlights the background region instead of the man's hand. 
Consequently, the model, which presumably has a bias towards an unfolded umbrella rather than a folded, hand-held umbrella,  generates a new person holding a red umbrella in the masked area instead of modifying the original subject as intended.

\noindent \textbf{Unfaithful Inversion for Real Video Data} \\
Although we demonstrated the applicability of our method to real-world videos in~\Cref{fig:real}, we found that existing open-sourced inversion methods for video generation models are not sufficiently faithful. 
In particular, they often fail to reconstruct latent key and value representations that are reliable enough for effective downstream editing. 
We believe this limitation could be mitigated with the development and public release of more robust inversion techniques tailored for video generation models.

A more in-depth analysis of these limitations is left to future work, as it falls beyond the scope of this study.

% \noindent \textbf{Does Not Guarantee Identity Preservation in Non-Rigid Editing}\\ 
% While key-value injection into non-vital layers enables impressive non-rigid editing, it does not guarantee identity preservation. We hypothesize that the global structure of the video is primarily encoded in vital layers, whereas non-vital layers mainly contribute to appearance information. Although high appearance resemblance often leads to satisfactory visual results, it does not necessarily preserve the exact identity of the object. For example, in~\Cref{fig:qual_all} (b), the shape of the turtle's shell differs from that in the source video, despite achieving a high degree of visual similarity.
\end{document}